\documentclass{article}

    \usepackage[nonatbib,preprint]{neurips_2020}

\usepackage[utf8]{inputenc} % allow utf-8 input
\usepackage[T1]{fontenc}    % use 8-bit T1 fonts
\usepackage{url, caption, subcaption}            % simple URL typesetting
\usepackage{booktabs, appendix}       % professional-quality tables
\usepackage{amsfonts}       % blackboard math symbols
\usepackage{nicefrac}       % compact symbols for 1/2, etc.
\usepackage{microtype, xcolor, soul}      % microtypography
\usepackage[ruled]{algorithm2e}
\usepackage{amsmath,amssymb,latexsym,amsthm}
\usepackage{graphicx,psfrag,epsf, csquotes,float}
\usepackage{wrapfig}

\definecolor{lightblue}{rgb}{0.188, 0.478, 0.858}
\definecolor{brown}{rgb}{0.76,0.5,0.18}
\newcommand{\refC}[1]{\textcolor{red}{\ref{#1}}}
\newcommand{\Ccite}[1]{\textcolor{lightblue}{\cite{#1}}}
\newcommand{\algo}[1]{\textcolor{brown}{\selectfont\ttfamily{{#1}}}}

\newtheorem{theorem}{Theorem}

\newtheorem{lemma}{Lemma}

\title{Differentiable Linear Bandit Algorithm}

\author{%
     Kaige Yang\\
    University College London\\
    \texttt{kaige.yang.11@ucl.ac.uk}\\
    \And 
    Laura Toni \\
    University College London\\
    \texttt{l.toni@ucl.ac.uk}\\
}

\begin{document}

\maketitle

\begin{abstract}
Upper Confidence Bound (UCB) is among the most commonly used methods for linear  multi-arm bandit problems. While  conceptually and computationally  simple,  this method highly relies on the confidence bounds, failing to  strike   the optimal exploration-exploitation if these bounds are not properly set. In the literature, confidence bounds are typically derived from concentration inequalities based on assumptions on the reward distribution, \textit{e.g.}, sub-Gaussianity.  The validity of these assumptions however is  unknown in practice.   
In this work, we aim at learning the confidence bound in a data-driven fashion, making it adaptive to the actual problem structure. Noting that existing UCB-typed algorithms are not differentiable with respect to confidence bound,   we first propose a novel differentiable linear bandit algorithm. Then, we  introduce a gradient estimator, which allows permit to learn the confidence bound   via iterative gradient ascent.   Theoretically, we show that the proposed algorithm achieves a $\tilde{\mathcal{O}}(\hat{\beta}\sqrt{dT})$ upper bound of $T$-round regret, where $d$ is the dimension of arm features and $\hat{\beta}$ is the learned size of the confidence bound.  Empirical results  show that $\hat{\beta}$ is significantly smaller than its theoretical upper bound and  proposed algorithms  outperform   baseline ones on both synthetic  and real-world datasets. 
\end{abstract}

\section{Introduction}
Multi-Arm Bandit (MAB) \Ccite{auer1995gambling} is an online decision making problem, in which an agent selects arms sequentially and observes stochastic rewards as feedback. The goal of the agent is to maximize the expected cumulative reward over a number of trials. The expected reward of each arm is unknown \textit{a priori} and it is learned from  experience by the agent. As a consequence, the agent needs to balance the selection of   arms to improve its  knowledge (exploration) and the selection of the highest rewarding arm given the knowledge acquired till thus far (exploitation). This is formalized as the so-called exploration-exploitation trade-off. Bandit algorithms are designed to strike this trade-off.
One class of MAB problems is the linear MAB \Ccite{chu2011contextual}, in which each arm is described by a feature vector and  the expected reward follows a linear model over its feature vector and an unknown parameter vector.
Each arm's feature vector is known  \textit{a priori} by the agent and it is considered as a hint on the arm reward. 
The learning problem boils down to the agent  inferring  the unknown parameter vector, based on the history (selected arms and received rewards) and selecting arms accordingly.

One popular algorithm to solve linear MAB is the \textit{Upper Confidence Bound} (UCB) \Ccite{auer2002using}  \Ccite{chu2011contextual} \Ccite{abbasi2011improved}. Its popularity is motivated by  its conceptual simplicity and strong theoretical guarantees. UCB-typed algorithms rely on the construction of an upper confidence bound, which is the estimated reward inflated based on the level of uncertainty of the estimate. At each decision opportunity, the agent selects the arm with the highest upper confidence bound. This reflects the \textit{Optimism in Face of Uncertainty}  principle. 
In such way, either the arm with high estimated reward (exploitation) or high uncertainty (exploration) is selected. However, to properly balance between exploration and exploitation, it is fundamental to establish a tight confidence bound~\Ccite{lattimore2016end}.

In most existing works, confidence bounds are derived from concentration inequalities \Ccite{abbasi2011improved} \Ccite{auer2002finite} \Ccite{mnih2008empirical} given \textit{a priori} assumptions on the reward distribution (e.g., sub-Gaussinaity). These bounds achieve strong minimax theoretical guarantees, outperforming competitor algorithms such as \algo{LinTS} \Ccite{agrawal2013thompson}.  While these bounds are essential for a theoretical analysis, they do not necessarily translate into practice.  In fact, these constructed confidence bounds are typically conservative \textit{in practice}, as noted in \Ccite{osband2015bootstrapped} \Ccite{kveton2018garbage}. 
This is because concentration inequalities are usually  built based on given reward distributions instead of the actual data (or problem structure). This results in non-adaptive and potentially wide confidence bounds which in turn lead to suboptimal performance in practice.

Alternatively, in this work we aim to learn the confidence bound in a data-driven fashion making it adaptive to the actual problem structure. Inspired by \Ccite{boutilier2020differentiable}, we aim at having a parametrized and differentiable cumulative reward function with respect to the confidence bound, which can then be optimized.  
The key challenge is that existing UCB-typed algorithms are  non-differentiable with respect to the confidence bound, mainly due to maximization of the UCB index (i.e., due to the presence of the $\arg\max$ operator in the \algo{OFUL} \Ccite{abbasi2011improved}, \algo{LinUCB} \Ccite{chu2011contextual}). To address this, we propose a novel differentiable UCB-typed linear bandit algorithm and introduce a gradient estimator which enables the confidence bound to be learned via gradient ascent.

Our proposed algorithm contains two core components. First, we consider a more informative UCB-based index than the classical UCB index used in \algo{OFUL} \Ccite{abbasi2011improved}, \algo{LinUCB} \Ccite{chu2011contextual}, which not only summarizes the history of each arm but also differentiates arms to be suboptimal arms and non-suboptimal arms. Second, we consider a softmax function, which transforms each  index into a probability distribution, where the probability for each suboptimal arm to be selected is arbitrary small. Conversely, the probability for a  non-suboptimal arm  to be selected is greater for arms with   larger index.
The key idea is that  the exploration is conducted by selecting arms with large index more often than others.
% \lt{the reader could argue that the index is like the UCB index that does both exploration and exploitation... I think we need to comment more on why this is mainly exploration..}
The exploitation is achieved by soft-eliminating suboptimal arms (arbitrary small probability to be selected). The softmax function ensures the differentiabilty of the reward function,  paving the way to learn confidence bound via gradient ascent.  Based on this, we provide two linear bandit algorithms for learning confidence bound in both  offline and online settings.  Theoretically, we provide a regret upper bound for the offline learning setting.

In summary, we make the following contributions:
\begin{itemize}
    \item We propose a novel UCB-typed linear bandit algorithm where the expected cumulative reward is a differentiable function of the confidence bound. 
    \item We introduce a gradient estimator and show how the confidence bound can be learned via gradient ascent both in offline/online settings.
    \item Theoretically, we prove a $\tilde{\mathcal{O}}(\hat{\beta} \sqrt{dT})$ upper bound of $T$-rounds regret where $\hat{\beta}$ is the learned size of the confidence bound in the offline setting.
    \item Empirically, we show $\hat{\beta}$ is significantly smaller than its theoretical upper bound,  leading to substantially lower cumulative regrets with respect to state-of-the-art baselines on synthetic and real-world datasets.
\end{itemize}

\textbf{Notation}:
$[K]$ mean the set $\{1,2,...,K\}$. Arm is indexed by $i, j \in \mathcal{A}$. We use boldface lower letter, e.g., $\mathbf{x}$, to denote vector and boldface upper letter. e.g., $\mathbf{M}$, to denote matrix. For a positive definite matrix $\mathbf{M}\in\mathbb{R}^{d\times d}$ and a vector $\mathbf{x}\in \mathbb{R}^d$, we denote the weighted 2-norm by $||\mathbf{x}||_{\mathbf{M}}=\sqrt{\mathbf{x}^T\mathbf{M}\mathbf{x}}$. Each arm $k$ is represented by the feature vector $\mathbf{x}_k \in \mathbb{R}^d$.
We denote by $\mathbb{P}$ and $\mathbb{E}$  the  probability distribution and the expectation operator, respectively.

%%%%%%%%%%%%%%%%%%%%%%%%%%%%%%%%%%%%%%%%%%%%%%%%%%%%
\section{Related work}
Our work is inspired by \Ccite{boutilier2020differentiable}, which was the first attempt in addressing policy-gradient optimization of bandit policies via differentiable bandit algorithm. However, there are fundamental differences between \Ccite{boutilier2020differentiable} and our work. First, authors  proposed a differential bandit framework for Bayesian MAB problem, which is not directly applicable to  linear MAB problems. Conversely,  we propose a differentiable UCB-typed linear bandit algorithm.
% \lt{do we want to add key things? which preserve ?? }
Second, the main goal of \Ccite{boutilier2020differentiable} is to learn 
% one of the inductive bias of statistical learning, specifically 
the learning rate (coldness-parameter) of the softmax function, while our algorithm aims at learning the size of the confidence bound.
%In relation to their work, we make the following contributions. First, their algorithm is not UCB-typed, while we propose a differential UCB-typed linear bandit algorithm. Second, our algorithms learn the size of confidence bound while their work learn the coldness-parameter (called learning rate therein) of the softmax function. 
Third, we propose algorithms for both offline and online settings, while  \Ccite{boutilier2020differentiable}   covered the offline setting only. Moreover,  
in \Ccite{boutilier2020differentiable} a regret analysis was provided for MAB with two arms. In contrast, we provide a regret analysis for linear MAB with arbitrary finite  number of arms in offline setting.

Another work focused on data-dependent UCB is \Ccite{hao2019bootstrapping}. Authors proposed an algorithm called \algo{bootstrapedUCB}. In \Ccite{hao2019bootstrapping}, the stochastic reward is assumed to be sub-Weibull random variable.  Multiplier bootstrap was employed to approximate the reward distribution. The boostrapped quantile acted as UCB to facilitie exploration. Their algorithm was deployed on both MAB and linear MAB problems, while regret analysis covered MAB only. Similar to this work, other bootstrap techniques were employed  \Ccite{chen2017ucb} \Ccite{elmachtoub2017practical} \Ccite{tang2015personalized}. 
Although aiming to the same goal (data-dependent UCB), these works are fundamentally differnt from our approach. Our algorithm is a differentiable bandit algorithm where we rely on gradient estimator to learn UCB. Their algorithm is non-differentiable, relying on the boostrapped quantile of the assumed reward distribution to construct UCB.

Bootstrap techniques were used also for Thompson Sampling exploration in \Ccite{osband2015bootstrapped}, in which author proposed the \algo{BoostrapThompson} algoritihm for MAB. Bootstrap techniques were used to sample observations from historical and pseudo observations to approximate the posterior distribution which was then used to encourage exploration.   As an extension, \Ccite{vaswani2018new} generalized this technique to Gaussian reward MAB, while \Ccite{kveton2019perturbed} and \Ccite{kveton2018garbage}  proposed an extension to contextual linear bandit, achieving  the same regret bound of \algo{LinTS} \Ccite{agrawal2013thompson}. The problem they aimed to address was the computational infeasibility  of inferring posterior distribution when reward follows nonlinear models. This departs from our goal, which is rather learning the confidence bound from data.

Our work can be viewed as a subtle combination of \algo{EXP3} \Ccite{auer1995gambling} and \algo{Phased Elimination}\footnote{Algorithm: \algo{Phased elimination with G-optimal exploration} page. 258 \Ccite{lattimore2018bandit}} \Ccite{lattimore2018bandit}.
\algo{EXP3} was designed for MAB, where arms with higher empirical averaged reward are signed with larger probability by softmax function.
The coldness-parameter of softmax function is a tunable hype-parameter chosen by the user. In our work, we propose a novel scheme to set this parameter automatically in a data-driven fashion. Moreover, although \algo{Exp3} is a differentiable bandit algorithm, it is not an UCB-typed algorithm. 
\algo{Phased Elimination} eliminates suboptimal arms based on the same index as ours and selects non-suboptimal arms uniformly (pure exploration). There are several fundamental differences between this approach and our work: \textit{i)} the confidence bound in our work is learned from data and not from concentration inequalities -- leading to a less conservative bound; \textit{ii)} \algo{Phased Elimination} is a non-differentiable algorithm; \textit{iii)} \algo{Phased Elimination} achieves  optimality in a worst case scenario (minmax regret) while our algorithm get an empirical gain being data dependent.

% is a phased-based algorithm which is designed to achieve the minimax optimal regret upper bound, while our algorithm in not phased-based \lt{maybe say why this is good?} and our goal is to learn confidence bound data-dependently instead of achieving minimax optimality.
% \lt{coudl we simplify saying that eliminatore g? }
% \blue{It is known that
% % The reason for this proportional probability being that
% selecting non-suboptimal arms uniformly at random results in unnecessary regret \Ccite{lattimore2018bandit}, failing in the goal of cumulative regret minimization.}
% \kg{This is tricky to say, since \algo{Successive elimination} achieves the minimax optimality, while performs poorly empirically.}

In summary, to the best of our knowledge, our work is the first differentiable UCB-typed linear bandit algorithm which enables confidence bound to be learned purely from data without relying on concentration inequalities and assumptions on the form of reward distribution.

\section{Problem setting}
\label{section: problem_setting}
We consider the stochastic linear bandit with 
an arm set $\mathcal{A}$ and a time horizon of $T$-rounds. The arm set contains $K$ arms, i.e., $|\mathcal{A}|=K$, where $K$ could be large. 
Each arm $i \in \mathcal{A}$ is associated with a known feature vector $\mathbf{x}_i \in \mathbb{R}^d$. The expected reward of each arm $\mu_i=\mathbf{x}_i^T\boldsymbol{\theta}$ follows a linear relationship over  $\mathbf{x}_i$ and an unknown parameter vector $\boldsymbol{\theta}$.
Similarly to other works in the  bandit literature, we assume that  arm feature and parameter vector are bounded $||\mathbf{x}||_2\leq L$ and $||\boldsymbol{\theta}||_2\leq C$, where $L>0$ and $C>0$. 
At the beginning of each decision opportunity $t\in [T]$, the learning agent selects one arm $i\in \mathcal{A}$ within the arm set $\mathcal{A}$. Upon this selection, the agent observes the instantaneous reward $y_t\in[0,1]$, which is  drawn independently from a distribution with unknown mean $\mu_i=\mathbf{x}_i^T\boldsymbol{\theta}$. 
% \lt{you could also expliclty say that it is $\mathbf{x}_i^T\boldsymbol{\theta}+n$ maybe?}\kg{since we make no assumption on noise $n$, I think no need to mention it.}
The agent aims to maximize the expected  cumulative reward over the time horizon $T$. Namely,
\begin{equation}
\label{eq: cum_reward}
     Y_T=\sum_{t=1}^T \mathbb{E}[y_t]
\end{equation}
This is equivalent to minimize the expected cumulative regret which measures the difference between the expected cumulative reward if the optimal arm were always selected and the agent's expected cumulative reward. Denoting   by $\mu_*=\max_{i \in \mathcal{A}}\mathbf{x}_i^T\boldsymbol{\theta}$   the expected reward of the optimal arm, we get  
\begin{equation}
\label{eq: cum_regret}
R_T=T\mu_*-\sum_{t=1}^T \mathbb{E}[y_t]\,.
\end{equation}
%Without loss of generality, we make the following assumption on the system model.  
%First, rewards are bound in $[0,1]$ for any arm $i$ and $t$ \lt{$y_t$?}. Second, arm feature and parameter vector are bounded $||\mathbf{x}||_2\leq L$ and $||\boldsymbol{\theta}||_2\leq C$, where $L>0$ and $C>0$. 

\textbf{Upper Confidence Bound (UCB)}. The upper confidence bound   algorithm, e.g., \algo{OFUL} \Ccite{abbasi2011improved}, is designed based on the   \textit{Optimism in Face of Uncertainty} principle. The key aspect is to construct a confidence bound of the estimated reward of each arm. Formally, at each round $t$, the confidence bound is defined as 
\begin{equation}
\label{eq: confidence_bound}
|\hat{\mu}_{i,t}-\mu_i|\leq \beta||\mathbf{x}_i||_{\mathbf{V}^{-1}_t}, \ \ \forall i\in \mathcal{A}
\end{equation}
where $\hat{\mu}_{i,t}$ is the estimate of the reward of arm $i$ at round $t$ and $\mathbf{V}_t=\sum_{t=1}^T \mathbf{x}_t \mathbf{x}_t^T$ is the Gram matrix up to round $t$.
Then, the agent selects the arm with the highest upper confidence bound as follows
\begin{equation}
\label{eq: ucb_index}
    i_t=\arg\max_{i \in \mathcal{A}}\hat{\mu}_{i,t}+\beta||\mathbf{x}_i||_{\mathbf{V}^{-1}_t}
\end{equation}
It is well known that the tighter the bound in Eq.~\refC{eq: confidence_bound}, the better the balance between exploration and exploitation \Ccite{lattimore2016end}. 
% \lt{if you have space you can also give an intuition on the importance of this --for example, a more conservative and less tight bounds would overestimate the uncertainty ...etc}
Most existing confidence bounds are established based on concentration inequalities. e.g., Hoeffding inequality \Ccite{auer2002finite}, self-normalized \Ccite{abbasi2011improved}, Azuma Inequality \Ccite{lattimore2018bandit}, Bernstein inequality \Ccite{mnih2008empirical}. As a specific example, under the assumption of the stochastic reward to be  a $R$-sub-Gaussian variable, one of the state-of-the-art high probability upper bound of $\beta$, derived based on properties of self-normalized martingale, was given by \Ccite{abbasi2011improved}:
\begin{equation}
    \label{eq: beta_upper_bound}
    \beta\leq R\sqrt{2\log\left(\frac{1}{\delta}\right)+d\log\left(1+\frac{T}{d}\right)}+\sqrt{\alpha}C
\end{equation}
where $\alpha$ is a regularizer parameter of least-square estimator, $1-\delta$ is the probability of which Eq.~\refC{eq: confidence_bound} holds and $||\boldsymbol{\theta}||_2\leq C$.
The tightness of this (and other bounds) relies on the validity of assumptions on the reward distribution, which is unfortunately unknown in practice.
Alternatively, we aim at learning the confidence bound, i.e., $\beta$, in a data-driven fashion without any \textit{a priori}  assumption on the unknown reward distribution except the linearity function of the  mean reward, i.e., is $\mu_i=\mathbf{x}_i^T\boldsymbol{\theta}, \forall i \in \mathcal{A}$.

\section{Algorithms}
In this section, we first present our proposed algorithm whose expected cumulative reward is a differentiable function of the confidence bound. Then, we provide a gradient estimator which enables confidence bound to be learned via gradient ascent. Next, we propose two algorithms to learn the confidence bound in offline and online settings, respectively. Finally, we prove a regret upper bound for offline learning setting.

\subsection{Differentiable Algorithm}
\begin{algorithm}[t]
\SetKwInOut{Input}{Input}\SetKwInOut{Output}{Output}
\Input{$\beta$, $\mathcal{A}$, $K$, $T$, $\alpha$. }
\BlankLine
\textbf{Initialization~~~:} $\mathbf{V}_0=\alpha \mathbf{I}\in \mathbb{R}^{d\times d}$, $\mathbf{b}_0=\mathbf{0}\in \mathbb{R}^d$, $\hat{\boldsymbol{\theta}}_0=\mathbf{0}\in \mathbb{R}^d$,  $\gamma_0=0$.\\
\For{$t \in [1, T]$}{
\begin{enumerate}
    \item Find $S_{i,t}, \forall i\in \mathcal{A}$ via Eq.~\refC{eq: s_i} with $\beta$.
    \item Find $\boldsymbol{\pi}_t$ via Eq.~\refC{eq: p_i} with $\gamma_{t-1}$.
    \item Select arm $i_t \in \mathcal{A}$ randomly following $\boldsymbol{\pi}_t$ and receive payoff $y_t$.
    \item Update $\mathbf{V}_t\gets \mathbf{V}_{t}+\mathbf{x}_t\mathbf{x}_t^T$, $\mathbf{b}_t\gets \mathbf{b}_{t-1}+\mathbf{x}_ty_t$ and $\hat{\boldsymbol{\theta}}_t=\mathbf{V}_{t}^{-1}\mathbf{b}_{t}$.
    \item Update $\gamma_t$ via Eq.~\refC{eq: emp_gamma} .
\end{enumerate}
}
\caption{\algo{SoftUCB}}
\label{algorithm: exp_ucb}
\end{algorithm}
Our proposed algorithm named \algo{SoftUCB} is shown in Algorithm~\refC{algorithm: exp_ucb}. 
\algo{SoftUCB} contains two core components: an UCB-based index $S_{i,t}$ and an arm selection policy $\boldsymbol{\pi}_t$. Formally, for $i\in \mathcal{A}$,  $\hat{\mu}_{i,t}=\mathbf{x}_i^T\hat{\boldsymbol{\theta}}_t$ where  $\hat{\boldsymbol{\theta}}_t=\mathbf{V}^{-1}_t\sum_{s=1}^t \mathbf{x}_s y_s$  is the least-square estimator and $\mathbf{V}^{-1}_t=\sum_{s=1}^t \mathbf{x}_s\mathbf{x}_s^T$ is the Gram matrix up to round $t$. Let denote by $i_*=\arg\max_{i \in \mathcal{A}}\hat{\mu}_{i,t}-\beta||\mathbf{x}_i||_{\mathbf{V}^{-1}_t}$ the arm with the largest lower confidence bound at round $t$.
Let us also  define  
\begin{align}
\phi_{i,t}  =||\mathbf{x}_i||_{\mathbf{V}^{-1}_t}+||\mathbf{x}_{i_*}||_{\mathbf{V}^{-1}_t}  \text{\ \  and \ \  }   \hat{\Delta}_{i,t}=\hat{\mu}_{i_*,t}-\hat{\mu}_{i,t} 
\end{align}
where  $\beta$ is the confidence bound  defined in Eq.~\refC{eq: confidence_bound} and $\hat{\Delta}_{i,t}$ is the estimated reward gap between $i_*$ and $i$.
Equipped with the above notations, we are now ready to introduce the  UCB-based index 
$S_{i,t}$  defined as 
\begin{equation}
\label{eq: s_i}
    S_{i,t}=\beta\phi_{i,t}-\hat{\Delta}_{i,t}\,.
\end{equation}
It is worth noting that $S_{i,t}$ is more informative than classical UCB index provided Eq.~\refC{eq: ucb_index}, because of the following two key properties: \textit{i)}, $S_{i,t}$ differentiates arms into suboptimal arms and non-suboptimal arms. Specifically, $S_{i,t}<0$ identifies arms which are suboptimal, $\Delta_i=\mu_*-\mu_i>0$, and therefore could be eliminated (i.e., not selected by the agent);
\textit{ii)}, $S_{i,t}\geq S_{j,t}\geq 0$ implies that  the upper confidence bound $\hat{\mu}_{i,t}+\beta|\mathbf{x}_i||_{\mathbf{V}^{-1}_t}\geq \hat{\mu}_{j,t}+\beta||\mathbf{x}_j||_{\mathbf{V}^{-1}_t}$ and therefore arm $i$ is more likely to be selected, in line with  the \textit{Optimism in Face of Uncertainty} principle. These two properties are stated formally in Lemma~\refC{lemma: suboptimal_arm}.
\begin{lemma}
\label{lemma: suboptimal_arm}
If $S_{i,t}<0$, arm $i$ is a suboptimal arm, i.e., $\mu_*-\mu_i>0$.  If $S_{i,t}\geq S_{j,t}\geq 0$, then the upper confidence bound  $\hat{\mu}_{i,t}+\beta||\mathbf{x}_i||_{\mathbf{V}^{-1}_t}\geq \hat{\mu}_{j,t}+\beta||\mathbf{x}_j||_{\mathbf{V}^{-1}_t}$. The proof is provided  in Appendix A.
\end{lemma}

We now describe the arm selection strategy. At each round $t \in [T]$, the probability for arm $i$ to  be selected is defined as 
\begin{equation}
\label{eq: p_i}
    p_{i,t}=\frac{\exp(\gamma_t S_{i,t})}{\sum_{j=1}^K \exp(\gamma_t S_{j,t})}
\end{equation}
where $\gamma_t>0$ is the coldness-parameter  controlling the concentration of the distribution (policy) $\boldsymbol{\pi}_{t}=[p_{1,t}, p_{2,t},...,p_{K,t}]$, 
and it is set as 
\begin{equation}
\label{eq: emp_gamma}
    \gamma_t=\frac{\log\left(\frac{\delta |\mathcal{L}_t|}{1-\delta}\right)}{\tilde{S}_{\text{max},t}}
\end{equation}
where at each round $t$, the arm set $\mathcal{A}$ is divided into two subsets $\mathcal{U}_t$ and $\mathcal{L}_t$ with $\mathcal{U}_t\cup \mathcal{L}_t=\mathcal{A}$ and $\mathcal{U}_t\cap \mathcal{L}_t=\emptyset$. Namely, $\mathcal{L}_t$ is the set of suboptimal arms (i.e., $i\in \mathcal{L}_t$ if $S_{i,t}<0$) and $\mathcal{U}_t$ is the set of non-suboptimal arms (i.e., $i\in \mathcal{U}_t$ if $S_{i,t}\geq 0$). $\tilde{S}_{\text{max},t}=\max_{i \in \mathcal{U}_t}S_{i,t}$,   $|\mathcal{L}_t|$ is the cardinality of $\mathcal{L}_t$ and 
$\delta$ is a probability hyper-parameter  explained in the following Lemma.   
\begin{lemma}
\label{lemma: gamma_lemma}
At any round $t\in [T]$, for any $\delta\in (0,1)$, setting $\gamma_t\geq \log(\frac{\delta|\mathcal{L}_t|}{1-\delta})/\tilde{S}_{\text{max},t}$ guarantees that $p_{\mathcal{U}_t}=\sum_{i\in \mathcal{U}_t}p_{i,t}\geq \delta $ and 
$p_{\mathcal{L}_t}=\sum_{i\in \mathcal{L}_t}p_{i,t}<1-\delta$. The proof is provided in Appendix B.
\end{lemma}

According to Lemma~\refC{lemma: gamma_lemma}, Eq.~\refC{eq: emp_gamma} guarantees that suboptimal arms ($i\in \mathcal{L}_t$) are selected with  an  arbitrary small probability (i.e., $p_{\mathcal{L}_t}<1-\delta \approx 0$ when $\delta \approx 1$). This leads to a soft-elimination of suboptimal arms. 
Furthermore, a positive $\gamma_t$ guarantees $p_{i,t}\geq p_{j,t}$ if $S_{i,t}\geq S_{j,t}\geq 0$, $\forall i, j\in \mathcal{U}_t$ which obeys the \textit{Optimism in Face of Uncertainty} principle.

Overall, \algo{SoftUCB} (soft-) eliminates suboptimal arms and selects non-suboptimal arms according to the index in  Eq.~\refC{eq: s_i} which favors the selection of arms with either high estimated reward or high uncertainty.

\subsection{Gradient Estimator of $\beta$}
We now show that the expected cumulative reward of \algo{SoftUCB} is a differentiable function over $\beta$ and introduce a gradient estimator. Formally, given the expected cumulative reward defined in Eq.~\refC{eq: cum_reward} and \algo{SoftUCB} described above, we have the optimization objective defined as
\begin{equation}
\label{eq: offline_objective}
\begin{split}
    \max_{\beta}Y_T=\max_{\beta}\sum_{t=1}^T\mathbb{E}[y_t]
    =\max_{\beta}\sum_{t=1}^T \sum_{i=1}^K p_{i,t}\mu_i,
    \ \ s.t. \  |\mu_i-\hat{\mu}_{i,t}|\leq \beta||\mathbf{x}_i||_{\mathbf{V}^{-1}_t}, \ \ \forall i\in \mathcal{A}, t\in [T]
\end{split}
\end{equation}
The imposed constraint  ensures that $\beta||\mathbf{x}_i||_{\mathbf{V}^{-1}_t}$ is indeed an actual   upper confidence bound (UCB) at any round $t \in [T]$ for any arm $i \in \mathcal{A}$. Applying the Lagrange multipliers gives the new objective:
\begin{equation}
\begin{split}
   \max_{\beta}\sum_{t=1}^T \sum_{i=1}^K p_{i,t}\mu_i-\eta (|\mu_i-\hat{\mu}_{i,t}|-\beta||\mathbf{x}_i||_{\mathbf{V}^{-1}_t}), \ \ s.t. \ \ \eta>0
\end{split}
\end{equation}
The gradient of $\beta$, denoted as $g(\beta)$, can be derived as (proof in Appendix C): 
\begin{equation}
\label{eq: beta_g}
   g(\beta)=\sum_{t=1}^T \sum_{i=1}^K p_{i,t}\mu_i \left(\gamma_t \phi_{i,t}-\frac{\sum_{j=1}^K \gamma_t \phi_{j,t}\exp (\gamma_t S_{j,t})}{\sum_{j=1}^K \exp (\gamma_t S_{j,t})}\right)+\eta||\mathbf{x}_i||_{\mathbf{V}^{-1}_t}
\end{equation}
Note that $\mu_i$ is unknown in practice and it is therefore replaced by its empirical estimate $\hat{\mu}_{i,t}$, leading to  the following gradient estimator  
\begin{equation}
\label{eq: emp_beta_g}
    \hat{g}(\beta)=\sum_{t=1}^T \sum_{i=1}^K p_{i,t}\hat{\mu}_{i,t} \left(\gamma_t \phi_{i,t}-\frac{\sum_{j=1}^K \gamma_t \phi_{j,t}\exp (\gamma_t S_{j,t})}{\sum_{j=1}^K \exp (\gamma_t S_{j,t})}\right)+\eta||\mathbf{x}_i||_{\mathbf{V}^{-1}_t}
\end{equation}
The gradient estimator $\hat{g}(\beta)$ in Eq.~\refC{eq: emp_beta_g} enables $\beta$ to be learned via gradient ascent. As a stochastic gradient method, under standard condition of learning rate, e.g., RM \Ccite{robbins1951stochastic}, it is expected that  $\hat{\beta}$ converges to local optimum.
\subsection{Training Settings}
\label{section: algorithms}
Equipped with the gradient estimator $\hat{g}(\beta)$(Eq.~\refC{eq: emp_beta_g}), we  now show how to learn $\beta$ in offline and online settings. The corresponding algorithms named \algo{SoftUCB offline} and \algo{SoftUCB online} are presented in Appendix E.

\textbf{Offline setting}. 
In this setting, multiple $T$-rounds trajectories of the bandit problem with the same arm set $\mathcal{A}$ are used to train $\beta$, which is refined  after each $T$-rounds trajectory.
The key steps are to initialize $\hat{\beta}_{0}$  and run \algo{SoftUCB} on $\mathcal{A}$ for $N$ training trajectories -- each trajectory containing $T$-rounds. After each trajectory $n\in [N]$,  update  $\hat{\beta}_n\gets \hat{\beta}_{n-1}+\lambda \hat{g}(\beta)$ via Eq.~\refC{eq: emp_beta_g} where $\lambda$ is the learning step. At the end of the  training, run \algo{SoftUCB} on $\mathcal{A}$ with $\hat{\beta}=\hat{\beta}_N$.

As a result of the training, the value of $\hat{\beta}$ is optimized in such a way that it maximizes the expected cumulative reward of arm set $\mathcal{A}$.
Empirically, the $\hat{\beta}$ to which the algorithm converges is substantial less than its theoretical upper bound Eq.~\refC{eq: beta_upper_bound}. This translates into a significant regret reduction.
In the following subsection we provide a  theoretical regret upper bound of \algo{SoftUCB offline}

While the above method is fully adaptive to the structure of $\mathcal{A}$, it provides a burden on the computational complexity. 
Specifically, the computational complexity \algo{SoftUCB offline} is $\mathcal{O}(NKT)$,  since we run \algo{SoftUCB} $N$ trajectories with $K$ arms and $T$ rounds in each trajectory.
This is much higher than other linear algorithms such as \algo{LinUCB}  \Ccite{abbasi2011improved} and \algo{LinTS} \Ccite{agrawal2013thompson}.  To mitigate this issue, we propose \algo{SoftUCB online} which learns $\beta$ within one trajectory in an online fashion.

\textbf{Online setting}. 
In this setting, $\hat{\beta}$ is updated online during one $T$-rounds trajectory. Specifically,   $\hat{\beta}_0$ is initialized  and  \algo{SoftUCB} on $\mathcal{A}$ is run for $T$ rounds. At the end of each round $t\in [T]$, update $\hat{\beta}_t\gets \hat{\beta}_{t-1}+\lambda \hat{g}_t(\beta)$ where $\lambda$ is the learning step and $\hat{g}_t(\beta)$ is the gradient estimator (Eq.~\refC{eq: emp_beta_g_online} defined blow). This reduces the computationally complexity to $\mathcal{O}(KT)$ since it does not require the $N$-training trajectories, which is at the same level of \algo{OFUL} \Ccite{abbasi2011improved}, \algo{LinUCB} \Ccite{chu2011contextual} and \algo{LinTS} \Ccite{agrawal2013thompson}.

In this setting, $Y_T=\sum_{t=1}^T \mathbb{E}[y_t]$, the objective function we aim at maximizing,  is not available before the end of the trajectory. To obviate to this problem, similarly to policy gradient methods for  non-episodic reinforcement learning problems~\Ccite{sutton2018reinforcement},  we update  $\hat{\beta}$ to maximizes the average reward per round $\hat{Y}_t$. 
Formally, at each round $t$,  $\hat{Y}_t$ consists of two parts: the observed  cumulative reward up to round $t$ and bootstrapped future reward under the current policy $\boldsymbol{\pi}_t=[p_{1,t}, p_{2,t},..., p_{K,t}]$. This translates in the following problem formulation
\begin{equation}
\label{eq: online_objective}
\begin{split}
    \max_{\beta}\hat{Y}_t
    &=\max_{\beta}\left(\sum_{s=1}^t\sum_{i=1}^K p_{i,s}\hat{\mu}_{i,s}+(T-t)\sum_{i=1}^K p_{i,t}\hat{\mu}_{i,t}\right)/T\\
    &s.t. \ \ |\hat{\mu}_{i,t}-\mu_{i,t}|\leq \beta||\mathbf{x}_i||_{\mathbf{V}_t^{-1}}, \forall i\in \mathcal{A}
\end{split}
\end{equation}
The gradient estimator $\hat{g}_t(\beta)$ at round $t$ can be derived as
\begin{equation}
\label{eq: emp_beta_g_online}
      \hat{g}_t(\beta)
      =\frac{1}{T}\bigg(\sum_{s=1}^t\sum_{i=1}^K \hat{\mu}_{i,s}\bigtriangledown_{\beta}p_{i,s}
      +(T-t)\sum_{i=1}^K \hat{\mu}_{i,t} \bigtriangledown_{\beta}p_{i,t}+\eta||\mathbf{x}_i||_{\mathbf{V}_t^{-1}}\bigg)
\end{equation}
It is worth noting that, at the end of trajectory $t=T$, the $\hat{Y}_t$ converges to. $Y_T$ in the offline setting.

\subsection{Theoretical Analysis}
\begin{theorem}
\label{theorem: lse_soft_regret}
Define $\mathbb{E}[r_t]=\mathbb{E}[\mu_*-\sum_{i=1}^K p_{i,t}\mu_i]$ be the expected regret at round $t\in [T]$.
Let $\hat{\beta}=\beta_N$ be  the confidence bound learned from the offline training setting  after $N$ $T$-rounds trajectories. Let assume that $\gamma_t$ follows Lemma~\refC{lemma: gamma_lemma} and $\delta \approx 1$. The cumulative regret of \algo{SoftUCB} is bounded as 
\begin{equation}
    R_T=\sum_{t=1}^T\mathbb{E}[r_t]\leq 4\sqrt{2}\hat{\beta} \delta \sqrt{Td\log\left(\alpha+\frac{T}{d}\right)}=\tilde{\mathcal{O}}\left(\hat{\beta} \sqrt{dT\log\left(1+\frac{T}{d}\right)}\right)
\end{equation}
where $\tilde{\mathcal{O}}(\cdot)$ hides absolute constant. The proof is contained in Appendix D. 
\end{theorem}
Theorem~\refC{theorem: lse_soft_regret} provides a regret upper bound of \algo{SoftUCB} in the offline setting. To compare the regret bound with that of other algorithms, we show $\hat{\beta}$ explicitly in the upper bound. Our regret bound scales with $d$ and $T$ as the regret bound $\mathcal{O}(\beta\sqrt{dT})$ of existing UCB-typed algorithms, e.g., \algo{OFUL} \Ccite{abbasi2011improved}, \algo{LinUCB} \Ccite{chu2011contextual}, \algo{Giro} \Ccite{kveton2019perturbed}. Since we make no assumption on the reward distribution, we can not derive a  theoretical upper bound on $\hat{\beta}$. However, it is worth to noting  that empirical results (in next section) show that $\hat{\beta}$ is significantly smaller than its theoretical upper bound Eq.~\refC{eq: beta_upper_bound}. The theoretical analysis for the online setting is left for future works.

\section{Experiments}
Our experimental evaluation aims to answer the following questions: (1) Does the learning curve of $\hat{\beta}$ converge in offline and online settings? (2) Is $\hat{\beta}$ lower than its theoretical counterpart? (3) How do our proposed algorithms  perform   compare to   baseline ones? 

\begin{table}[t]
  \caption{The comparison between $\hat{\beta}$ (offline) and theoretical bound $\tilde{\beta}$}.
  \label{sample-table}
  \centering
  \begin{tabular}{lllll}
    \toprule
    $d=5, T=2^{8}$ & $d=5, T=2^{9}$ & $d=5, T=2^{10}$ & $d=10, T=2^{10}$ & $d=15, T=2^{10}$ \\
    \midrule
    $\hat{\beta}=\textbf{0.5}$  &  $\hat{\beta}=\textbf{0.6}$ &  $\hat{\beta}=\textbf{0.9}$ & $\hat{\beta}=\textbf{1.1}$ & $\hat{\beta}=\textbf{1.2}$   \\
    $\tilde{\beta}=2.56$  &  $\tilde{\beta}=2.66$ &  $\tilde{\beta}=2.76$ & $\tilde{\beta}=3.25$ & $\tilde{\beta}=3.61$  \\
    \bottomrule
  \end{tabular}
\label{table: beta_vs_beta_hat}
\end{table}

\begin{figure}[t]
    \centering
    \begin{subfigure}{0.24\textwidth}
        \includegraphics[width=\textwidth, height=4cm]{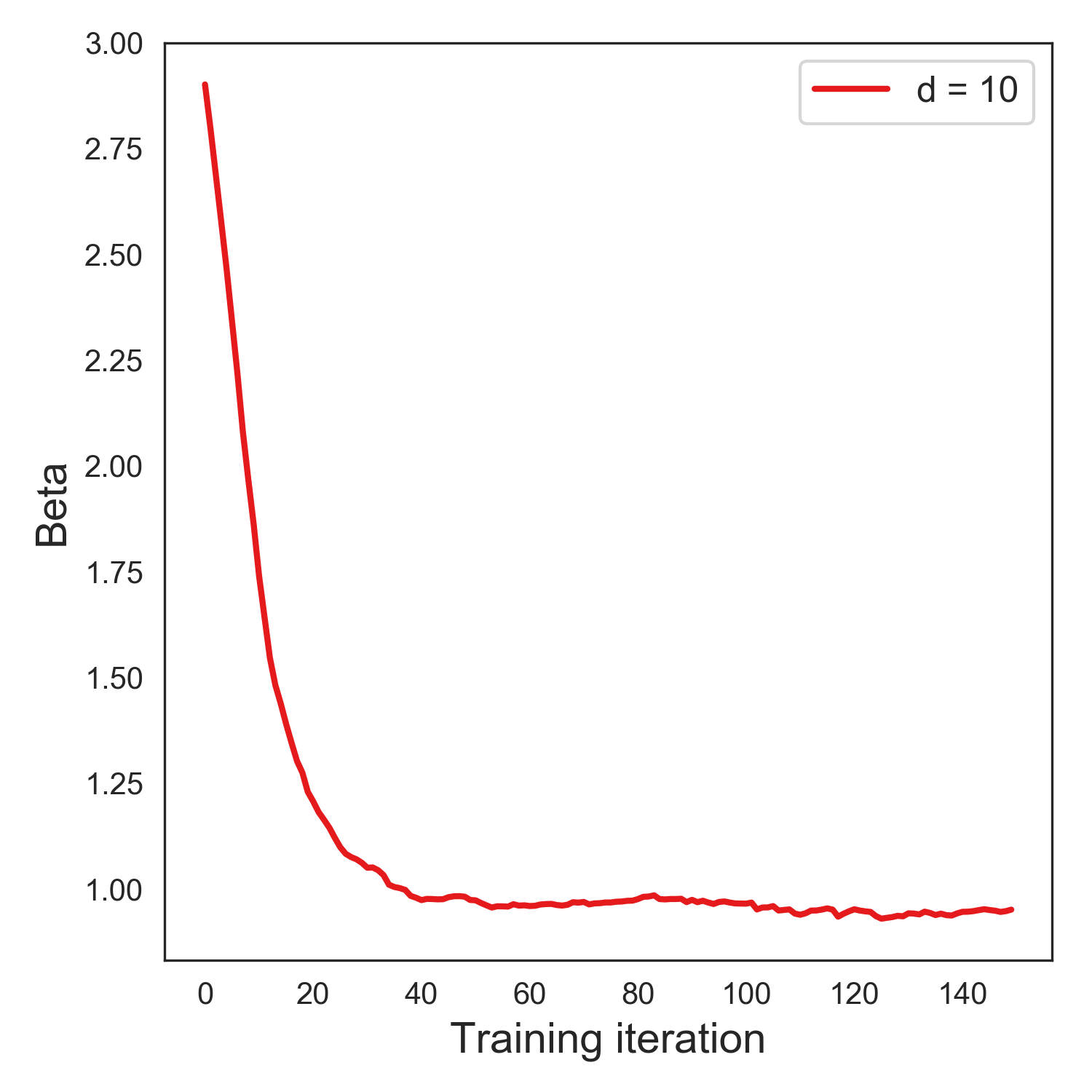}
        \subcaption{$\hat{\beta}$ (offline)} 
        \label{fig: learn_curves_beta_offline}
    \end{subfigure}
    \begin{subfigure}{0.24\textwidth}
        \includegraphics[width=\textwidth, height=4cm]{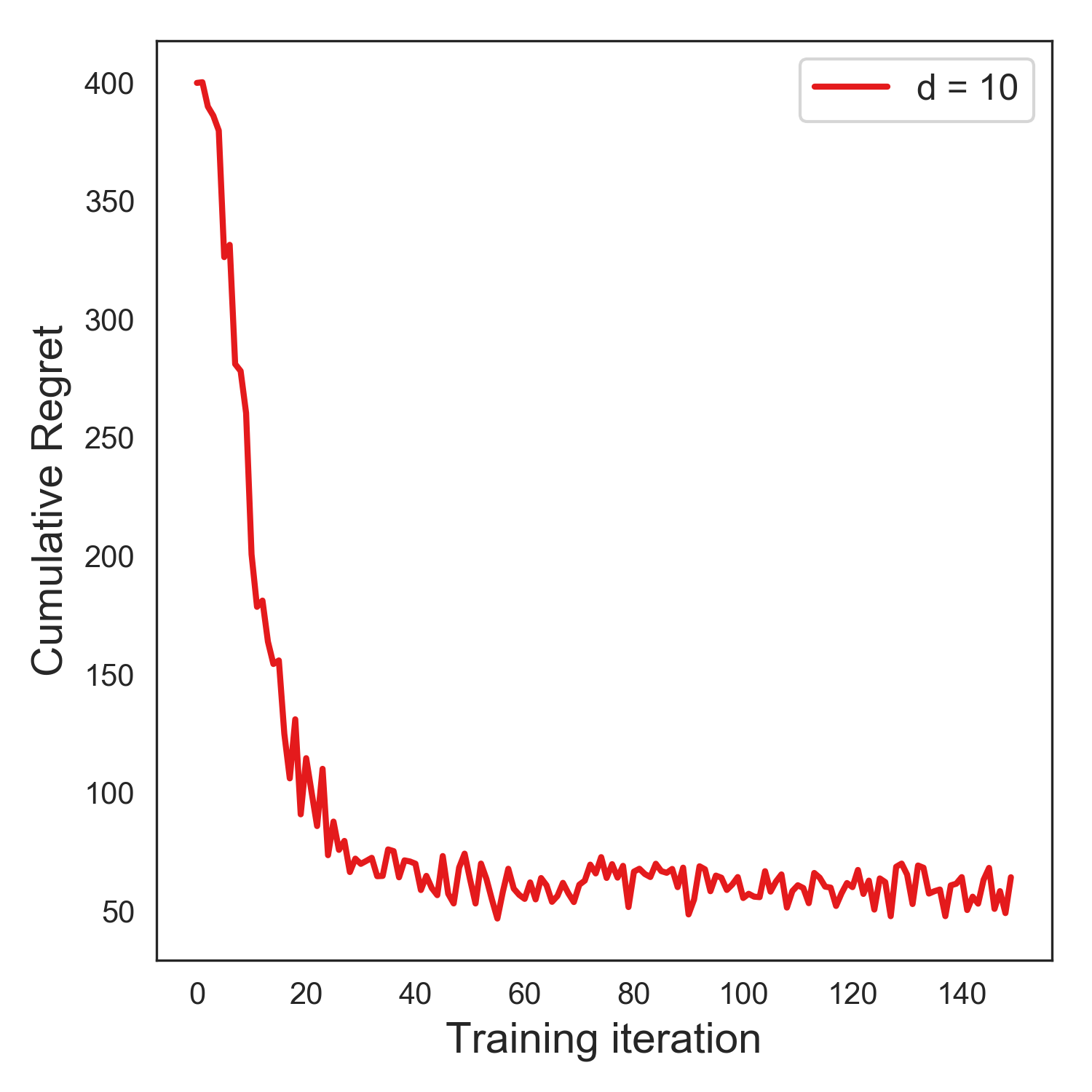}
        \subcaption{$R_T$ (offline)}
        \label{fig: learn_curves_regret_offline}
    \end{subfigure}
    \begin{subfigure}{0.24\textwidth}
        \includegraphics[width=\textwidth, height=4cm]{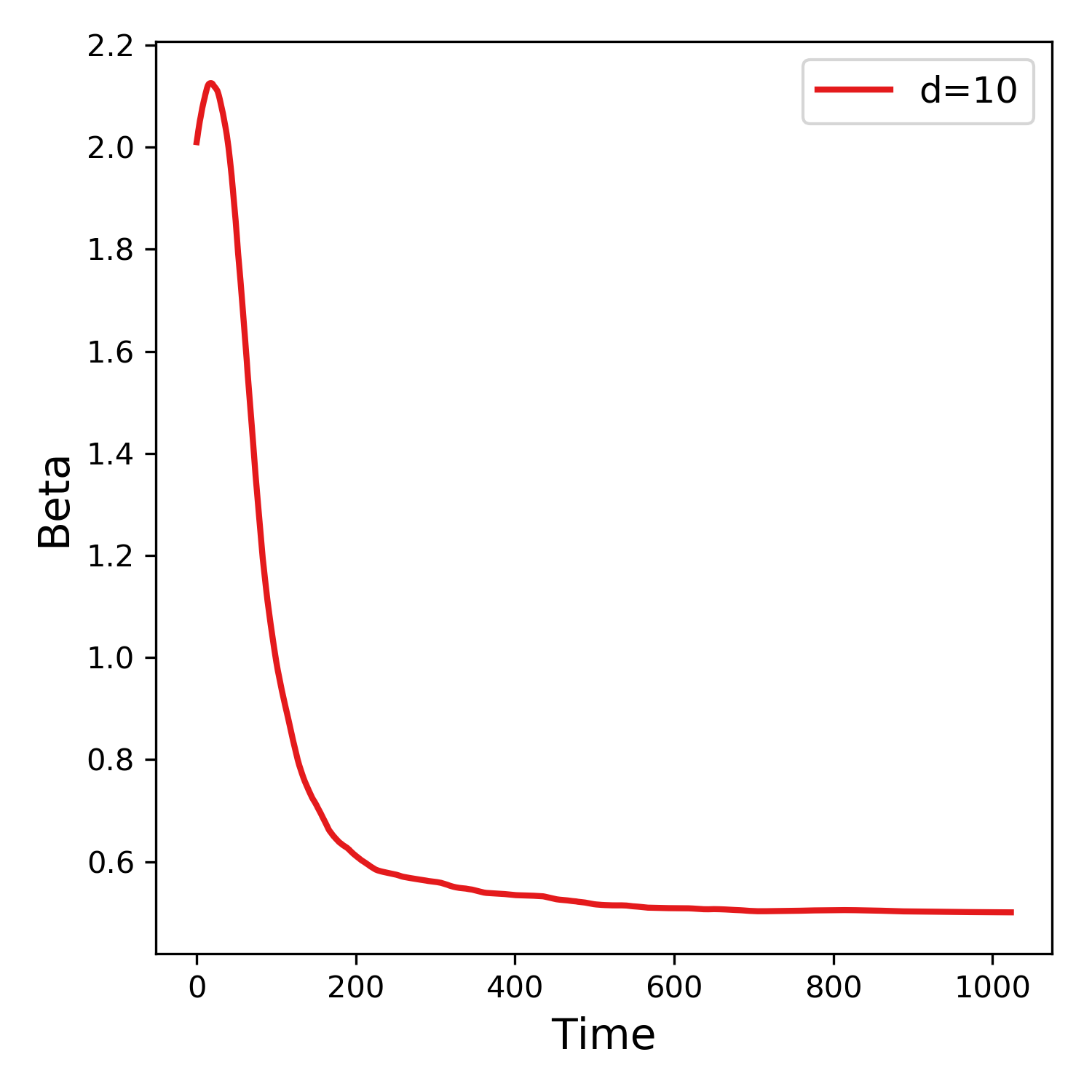}
        \subcaption{$\hat{\beta}$ (online)}
        \label{fig: learn_curves_beta_online}
    \end{subfigure}
    \begin{subfigure}{0.24\textwidth}
        \includegraphics[width=\textwidth, height=4cm]{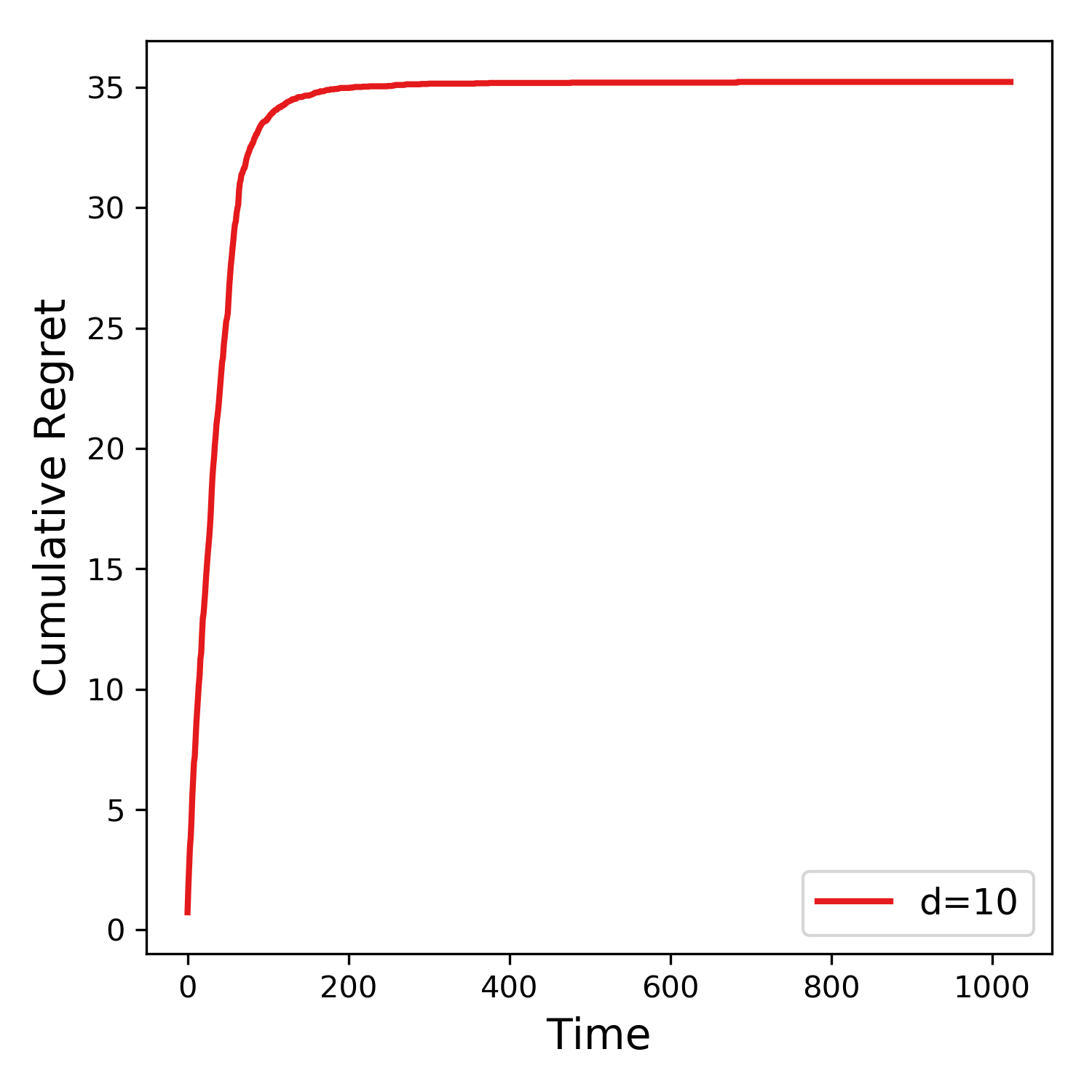}
        \subcaption{$R_T$ (online)}
        \label{fig: learn_curves_regret_online}
    \end{subfigure}
\caption{Learning curves of \algo{SoftUCB offline} and \algo{SoftUCB online}}
\label{fig: learn_curves}
\end{figure}

\begin{figure}[t]
    \centering
    \begin{subfigure}{0.24\textwidth}
        \includegraphics[width=\textwidth,height=4cm]{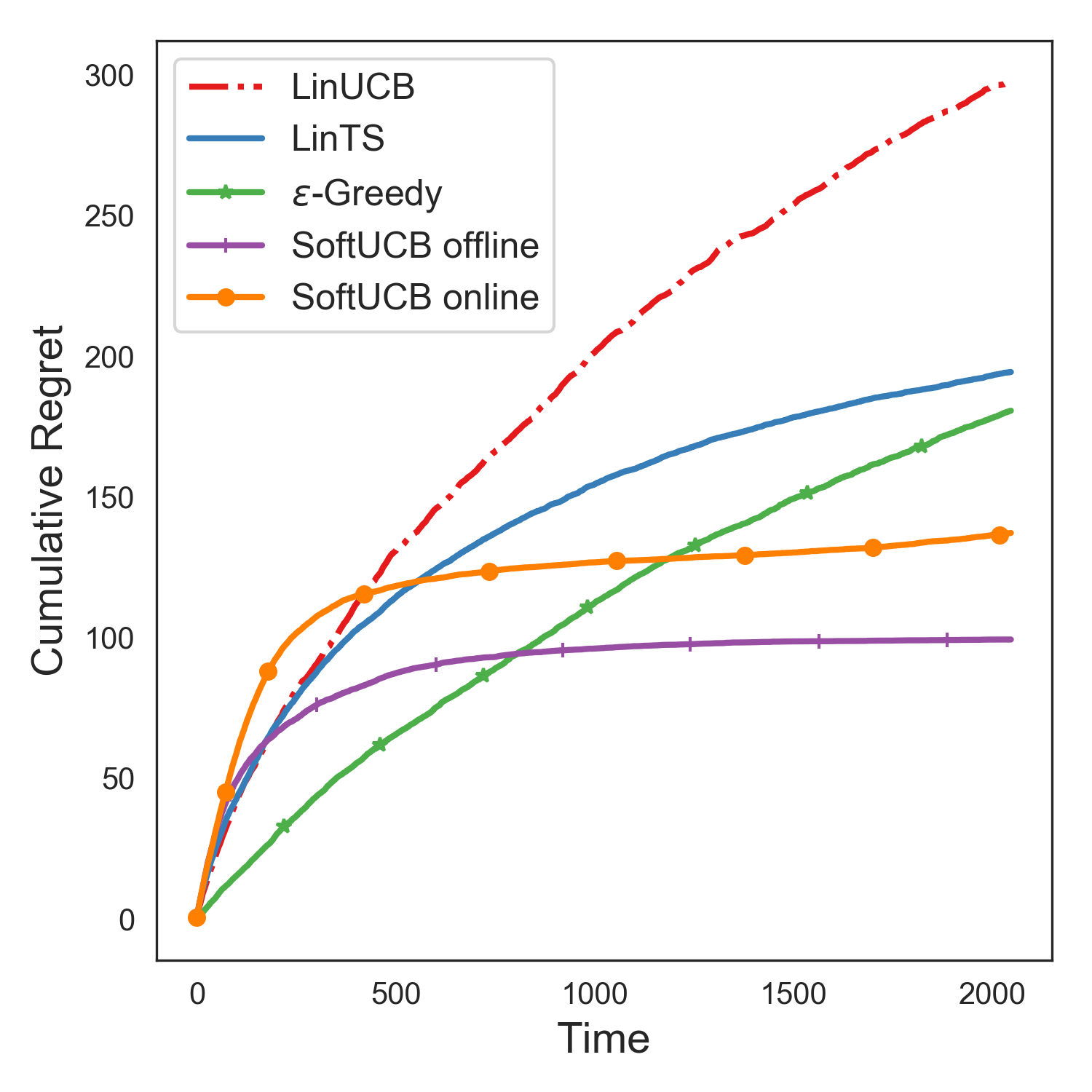}
        \subcaption{$d=10$}
        \label{fig: simu_d_10}
    \end{subfigure}
    \begin{subfigure}{0.24\textwidth}
        \includegraphics[width=\textwidth,height=4cm]{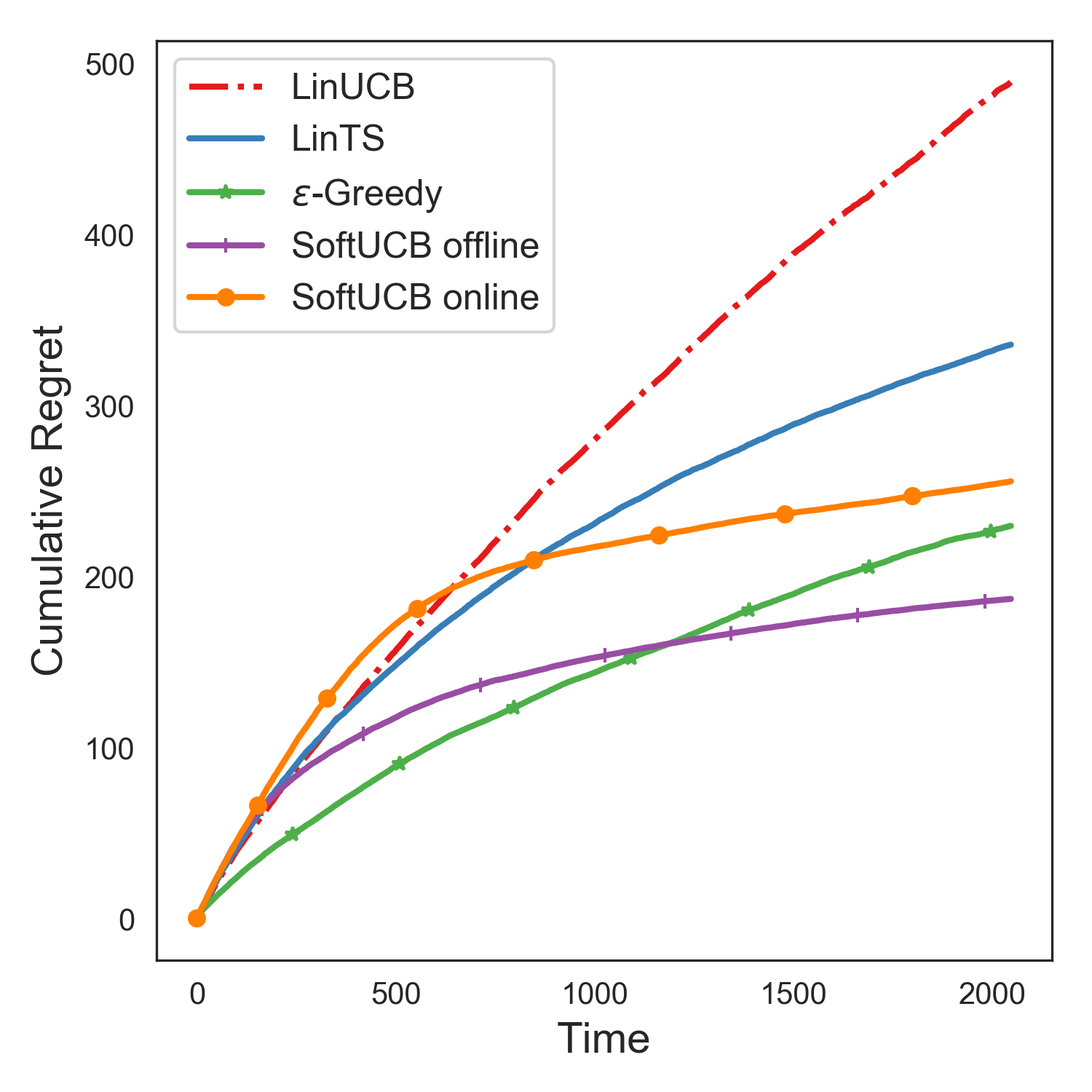}
        \subcaption{$d=20$}
        \label{fig: simu_d_20}
    \end{subfigure}
    \begin{subfigure}{0.24\textwidth}
        \includegraphics[width=\textwidth,height=4cm]{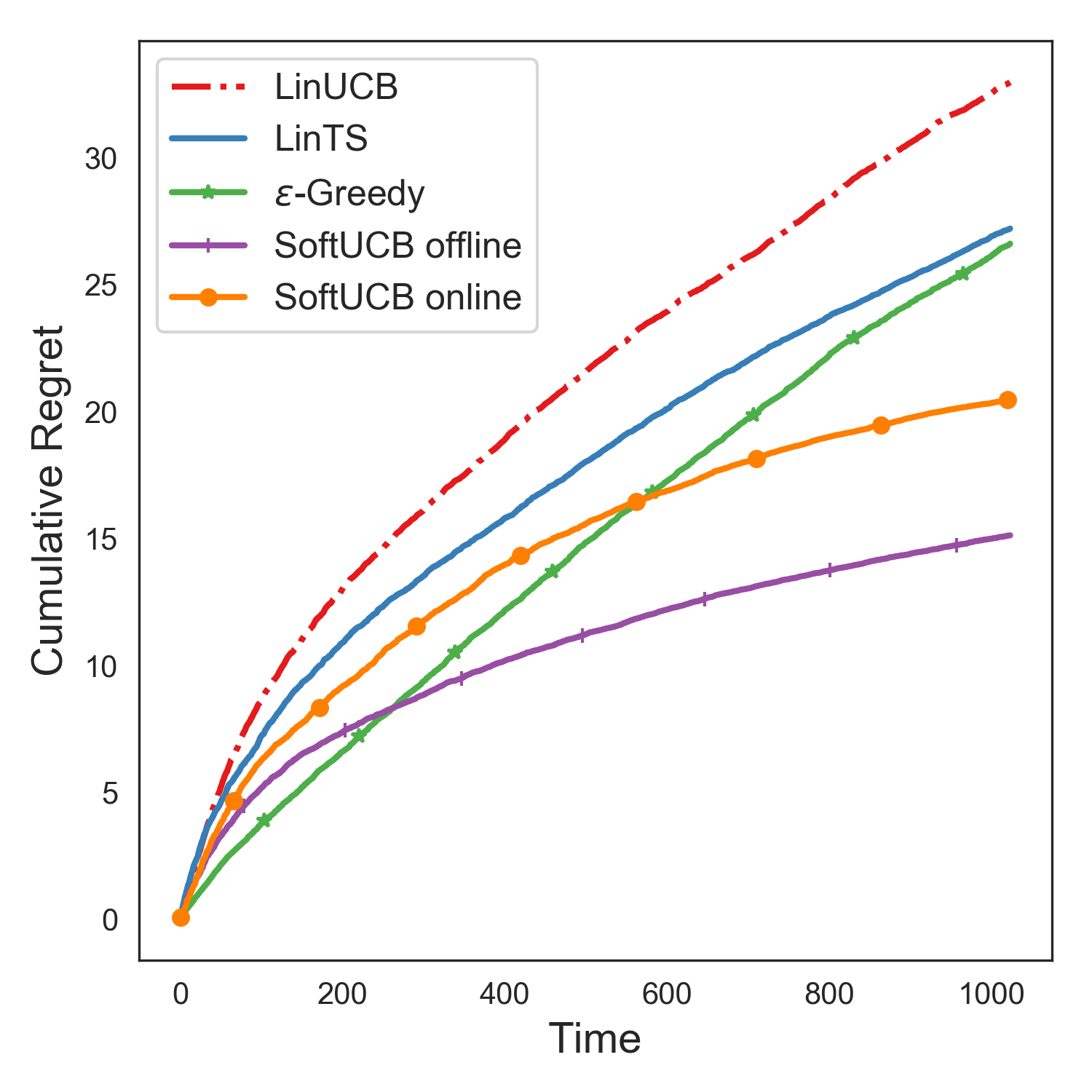}
        \subcaption{\textbf{MovieLens}}
        \label{fig: movie}
    \end{subfigure}
    \begin{subfigure}{0.24\textwidth}
        \includegraphics[width=\textwidth,height=4cm]{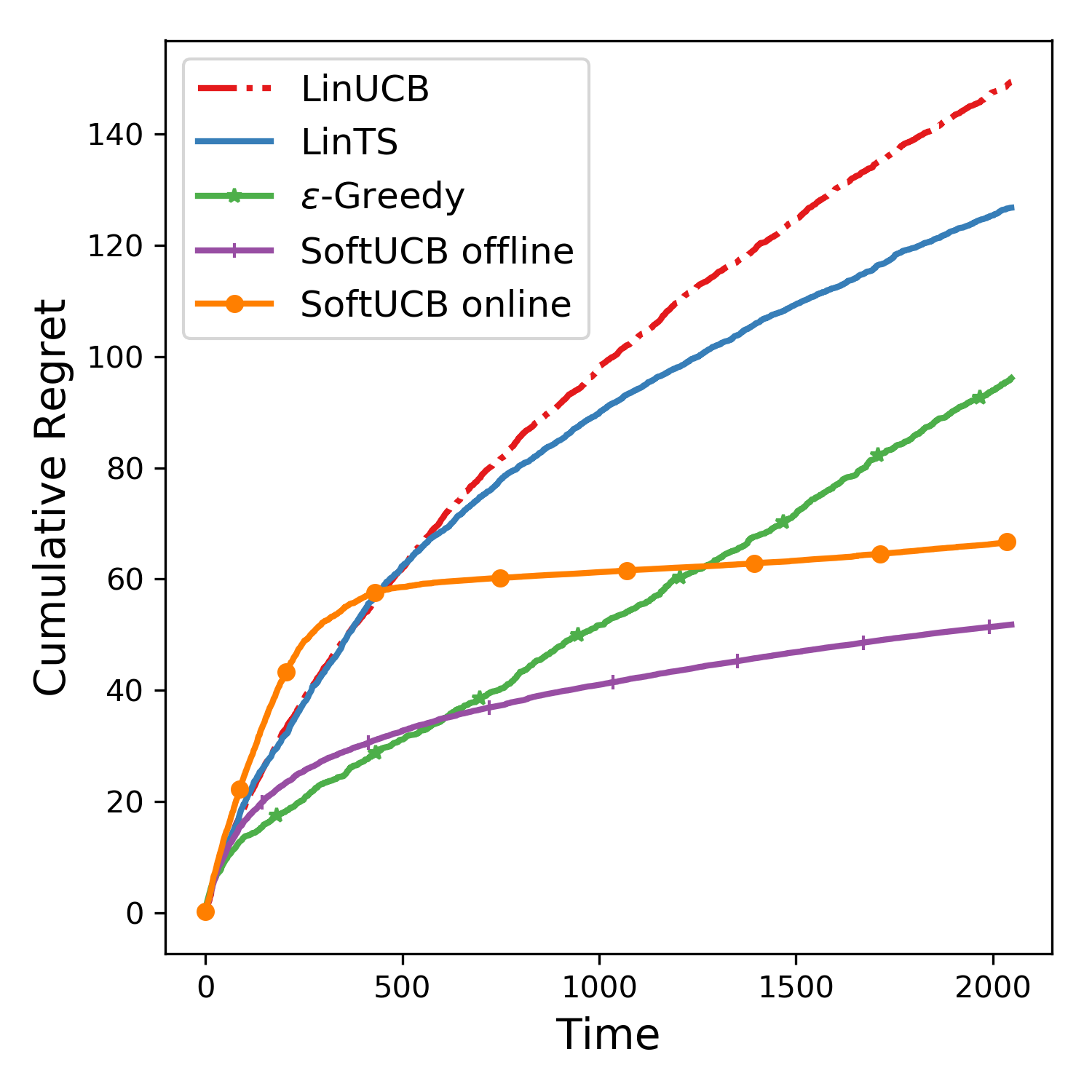}
        \subcaption{\textbf{Jester}}
        \label{fig: jester}
    \end{subfigure}
\caption{Performance of algorithms on synthetic and real-world datasets}
\label{fig: perf_baseline}
\end{figure}

In synthetic datasets, there are $K=50$ arms with feature vector drawn uniformly from $[-1, 1]$. The dimension of arm feature is set as $d=10,20$. Arm feature vectors are normalized to be unit vectors. The parameter vector $\boldsymbol{\theta}$ is generated as a random unit vector. The noise level is set as $0.5$ and the regularizer parameter  is $\alpha=1$. 
We use two real-world datasets: \textbf{Jester} \Ccite{goldberg2001eigentaste} and \textbf{Movielens} \Ccite{lam2006movielens} (see  Appendix G for more details).
We compare the proposed algorithms with baseline ones, namely \algo{LinUCB} \Ccite{abbasi2011improved}, \algo{LinTS} \Ccite{agrawal2013thompson} and \algo{$\epsilon$-greedy} \Ccite{sutton2018reinforcement}. The $\beta$  in \algo{LinUCB} is set as Eq.~\refC{eq: beta_upper_bound}, \algo{LinTS} follows \Ccite{agrawal2013thompson}, and $\epsilon=0.05$ in \algo{$\epsilon$-greedy}.

Fig.~\refC{fig: learn_curves} depicts the learning curves of $\hat{\beta}$ and the corresponding $R_T$ in both offline and online settings for the synthetic datasets. The feature dimension $d=10$.
In both settings, $\hat{\beta}$ and $R_T$ achieve convergence. Note that in offline setting, $\hat{\beta}$ is optimized to maxmize the expected cumulative reward Eq.~\refC{eq: offline_objective}, while in online setting, $\hat{\beta}$ is optimized to maximize the average reward per round Eq.~\refC{eq: online_objective}.

In Table~\refC{table: beta_vs_beta_hat}, we compare $\hat{\beta}$ obtained from offline training and its theoretical suggested $\tilde{\beta}$ given by Eq.~\refC{eq: beta_upper_bound}. Clearly, $\hat{\beta}$ is significantly less than $\tilde{\beta}$ consistently in all cases. This is because $\hat{\beta}$ is adaptive to the structure of $\mathcal{A}$, while $\tilde{\beta}$ is derived based on worst-case (minimax analysis). Note that the value of $\hat{\beta}$ is highly data-dependent. The value we report here only valid for our experimental data. However, it is reasonable to expect $\hat{\beta}$ less that $\tilde{\beta}$ in general. The corresponding learning curves are shown in Appendix F.

In Fig.~\refC{fig: perf_baseline}
, the proposed algorithms converge to lower cumulative regret comparing with baselines. There are two reasons: First, the confidence bound $\hat{\beta}$ is optimized. Second, the proposed algorithm eliminates (softly) suboptimal arms which accelerates the rate of convergence. It is worth noting that the regret of \algo{SoftUCB online} is large at the initial phase. This is because at the beginning, when $\gamma_0=0, |\mathcal{L}_t|=0$, \algo{SoftUCB online} selects arms uniformly which results in large regret. Later, when suboptimal arms are identified, $|\mathcal{L}_t|>0$, $\gamma_t>0$ according to Eq.~\refC{eq: emp_gamma}. Suboptimal arms are soft-eliminated and non-suboptimal arms are selected following index Eq.~\refC{eq: s_i} which controls the regret.

Finally, during our experiments, we noticed that the convergence of $\hat{\beta}$ in both offline and online setting is sensitive to the Lagrange multiplier $\eta$. With large $\eta$, the gradient ascent algorithm fails in converging, this is because the gradient estimator Eq.~\refC{eq: emp_beta_g_online} is dominated by $\eta||\mathbf{x}_i||_{\mathbf{V}_{t}^{-1}}$. On the other hand, too small $\eta$ does not ensure the key constraint  $|\hat{\mu}_{i,t}-\mu_{i}|\leq \beta||\mathbf{x}_i||_{\mathbf{V}^{-1}_t}$. This can lead to erroneously eliminating  the optimal arm. Therefore, the hyper-parameter $\eta$ needs to be tuned carefully during experiments.

\section{Conclusion}
We propose \algo{SoftUCB}, a novel UCB-typed linear bandit algorithm based on an \textit{adaptive} confidence bound, resulting in a less conservative algorithm respect to  UCB-typed algorithms with \textit{constructed} confidence bounds.  The key novelty is to propose an expected cumulative reward which  is a differentiable function of the confidence bound, and derive  a gradient estimator, which enables confidence bound to be learned via gradient ascent. 
The estimated confidence bound $\hat{\beta}$ can be updated under offline/online training settings with the proposed    \algo{SoftUCB offline} and \algo{SoftUCB online}, respectively.
Theoretically, we provide a $\tilde{\mathcal{O}}(\hat{\beta} \sqrt{dT})$ regret upper bound of \algo{SoftUCB} in the offline setting. Empirically, we show that   $\hat{\beta}$ is significantly  less that its theoretical counterpart leading to a reduction of the cumulative regret  compared to  state-of-the-art baselines.

There are several directions for future work. First, our work can be combined with meta-learning algorithms, e.g., \algo{MAML} \Ccite{finn2017model}, to learn a confidence bound which is adaptive to the common structure of a set of bandit tasks. Second, we believe our work can be generalized to reinforcement learning (RL) tasks where exploration and exploitation trade-off is a long standing challenge.

\newpage
\section{Broader Impact Discussion}
Our work is an algorithm for multi-arm bandit (MAB) problem. On the novelty side, our work automates the exploration in bandit problems.
Such algorithm could be used in recommendation system and clinic trials. On the positive side, our work could balance the exploration and exploitation trade-off in a problem dependently way, which might improve the customer satisfaction or patient's health care. On the negative side,
depending to the deployed application, the recommended contents might be unsuitable for some users. To mitigate this issue, domain knowledge might be required to filter the recommended contents before releasing to users. Regarding the health care application, expert's supervision is essential to avoid any potential hazard.
\bibliographystyle{plain}
\bibliography{neurips_2020}

\newpage

\section*{Appendidx A}
This section contains the proof of Lemma~\refC{lemma: suboptimal_arm}.
\begin{proof}
Suppose $S_{i,t}<0$, that is
\begin{equation}
    \beta(||\mathbf{x}_{i_*}||_{\mathbf{V}^{-1}_t}+||\mathbf{x}_i||_{\mathbf{V}^{-1}_t})<\hat{\mu}_{i_*,t}-\hat{\mu}_{i,t}
\end{equation}
Rearrange terms gives
\begin{equation}
    \hat{\mu}_{i,t}+\beta||\mathbf{x}_i||_{\mathbf{V}^{-1}_t}\leq \hat{\mu}_{i_*,t}-\beta||\mathbf{x}_{i_*}||_{\mathbf{V}^{-1}_t}
\end{equation}
Note that $|\mu_i-\hat{\mu}_{i,t}|\leq \beta||\mathbf{x}_i||_{\mathbf{V}^{-1}_t}, \ \forall i\in \mathcal{A}$. Then, 
\begin{equation}
    \hat{\mu}_{i_*,t}-\beta||\mathbf{x}_{i_*}||_{\mathbf{V}^{-1}_t}\leq \mu_{i_*}
\end{equation}
and 
\begin{equation}
    \mu_i\leq \hat{\mu}_{i,t}+\beta||\mathbf{x}_i||_{\mathbf{V}^{-1}_t}
\end{equation}
Combine together we have
\begin{equation}
    \mu_i\leq \mu_{i_*}\leq \mu_*
\end{equation}
Recall by definition $i_*=\arg\max_{i\in \mathcal{A}}\hat{\mu}_{i,t}-\beta||\mathbf{x}_i||_{\mathbf{V}_t^{-1}}$ is the arm with largest lower upper bound at round $t$.
Therefore, $\Delta_i=\mu_*-\mu_i>0$. In words, arm $i$ is suboptimal.\\
Suppose $S_{j,t}\geq S_{i,t}\geq 0$ 
\begin{equation}
    \beta(||\mathbf{x}_{j_*}||_{\mathbf{V}^{-1}_t}+||\mathbf{x}_j||_{\mathbf{V}^{-1}_t})-(\hat{\mu}_{j_*,t}-\hat{\mu}_{j,t})\leq \beta(||\mathbf{x}_{i_*}||_{\mathbf{V}^{-1}_t}+||\mathbf{x}_i||_{\mathbf{V}^{-1}_t})-(\hat{\mu}_{i_*,t}-\hat{\mu}_{i,t})
\end{equation}
Recall the definition of $i_*$, 
\begin{equation}
    i_*=\arg \max_{j \in [K]}\hat{\mu}_{j,t}-\beta||\mathbf{x}_j||_{\mathbf{V}^{-1}_t}
\end{equation}
Thus, at each time $t$, $i_*=j_*$. Then,
\begin{equation}
\beta||\mathbf{x}_j||_{\mathbf{V}^{-1}_t}+\hat{\mu}_{j,t}\leq \beta||\mathbf{x}_i||_{\mathbf{V}^{-1}_t}+\hat{\mu}_{i,t}
\end{equation}
\end{proof}

\section*{Appendix B}
This section contains the proof of Lemma~\refC{lemma: gamma_lemma}.
\begin{proof}
\begin{equation}
    p_{\mathcal{U}_t}=\frac{\sum_{i\in \mathcal{U}_t}\exp(\gamma_t S_{i,t})}{\sum_{i\in \mathcal{U}_t}\exp(\gamma_t S_{i,t})+\sum_{j\in \mathcal{L}_t}\exp(\gamma_t S_{j,t})}
\end{equation}
By definition, $S_{j,t}<0$, $\forall j\in \mathcal{L}$. Thus, 
\begin{equation}
    \exp(\gamma S_{j,t})< 1, \ \forall j \in \mathcal{L}
\end{equation}
Then, 
\begin{equation}
    \sum_{j\in \mathcal{L}_t}\exp(\gamma S_{j,t})< |\mathcal{L}_t| 
\end{equation}
Therefore, 
\begin{equation}
    p_{\mathcal{U}_t}>\frac{\sum_{i\in \mathcal{U}_t}\exp(\gamma_t S_{i,t})}{\sum_{i\in \mathcal{U}_t}\exp(\gamma_t S_{i,t})+|\mathcal{L}_t|}
\end{equation}
For any probability $\delta\in (0,1)$, we can find a $\gamma_t$ such that $p_{\mathcal{U}_t}\geq \delta$, namely
\begin{equation}
    \frac{\sum_{i\in \mathcal{U}_t}\exp(\gamma_t S_{i,t})}{\sum_{i\in \mathcal{U}_t}\exp(\gamma_t S_{i,t})+|\mathcal{L}_t|}\geq\delta
\end{equation}
Rearrange terms gives
\begin{equation}
\label{eq: w_bound}
    \sum_{i\in \mathcal{U}_t}\exp(\gamma_t S_{i,t})\geq\frac{\delta |\mathcal{L}_t|}{1-\delta}
\end{equation}
Take logarithm on both sides, 
\begin{equation}
    \log\left(\sum_{i\in \mathcal{U}_t}\exp(\gamma_t S_{i,t})\right)\geq\log\left(\frac{\delta|\mathcal{L}_t|}{1-\delta}\right)
\end{equation}
The left side term is LogSumExp which can be approximated by 
\begin{equation}
    \log\left(\sum_{i\in \mathcal{U}_t}\exp(\gamma_t S_{i,t})\right)\geq \max_{i \in \mathcal{U}_t}\gamma_t S_{i,t}=\gamma_t \max_{i\in \mathcal{U}_t}S_{i,t}
\end{equation}
Denote $\tilde{S}_{\text{max},t}=\max_{i\in \mathcal{U}_t}S_{i,t}$ and let 
\begin{equation}
    \gamma_t \tilde{S}_{\text{max},t}\geq\log(\frac{\delta|\mathcal{L}_t|}{1-\delta})
\end{equation}
we have
\begin{equation}
\label{eq: gamma_bound}
    \gamma_t\geq\frac{\log(\frac{\delta \mathcal{L}_t}{1-\delta})}{\tilde{S}_{\text{max},t}}
\end{equation}
Therefore, if $\gamma_t$ satisfies Eq.~\refC{eq: gamma_bound}, 
\begin{equation}
    p_{\mathcal{U}_t}\geq\delta
\end{equation}
Clearly, $p_{\mathcal{L}_t}<1-\delta$ since $p_{\mathcal{L_t}}+p_{\mathcal{U}_t}=1$.
\end{proof}

\section*{Appendix C}
This section contains the derive of gradients.
\begin{proof}
\begin{equation}
\begin{split}
    \max_{\beta}Y(T)
    &=\max_{\beta}\sum_{t=1}^T\mathbb{E}[y_t]=\max_{\beta, \gamma}\sum_{t=1}^T \sum_{i=1}^K p_{i,t}\mu_i\\
    &s.t. \ \  |\mu_i-\hat{\mu}_{i,t}|-\beta||\mathbf{x}_i||_{\mathbf{V}^{-1}_t}\leq 0, \ \ \forall i\in \mathcal{A}, \ \  \forall t\in [T]
\end{split}
\end{equation}
Apply the Lagrange multipliers, the optimization objective is 
\begin{equation}
\begin{split}
    \max_{\beta}\sum_{t=1}^T \sum_{i=1}^K p_{i,t}\mu_i-\eta (|\mu_i-\hat{\mu}_{i,t}|-\beta||\mathbf{x}_i||_{\mathbf{V}^{-1}_t})\ \ s.t. \ \ \eta>0
\end{split}
\end{equation}

Apply the score function $\bigtriangledown_\theta f(\theta)=f(\theta)\bigtriangledown_\theta \log f(\theta)$ to $p_{i,t}$\\
\begin{equation}
    \log p_{i,t}=\gamma S_{i,t}-\log \sum_{j=1}^K \exp \gamma S_{j,t}
\end{equation}
\begin{equation}
    \bigtriangledown_{\beta}\log p_{i,t}=\gamma_t \phi_{i,t}-\frac{\sum_{j=1}^K \gamma_t \phi_{j,t}\exp \gamma_t S_{j,t}}{\sum_{j=1}^K \exp \gamma_t S_{j,t}}
\end{equation}
Then, the gradient $g(\beta)$ is
\begin{equation}
    g(\beta)=\sum_{t=1}^T \sum_{i=1}^K \mu_i p_{i,t}\left(\gamma_t \phi_{i,t}-\frac{\sum_{j=1}^K \gamma_t \phi_{j,t}\exp \gamma_t S_{j,t}}{\sum_{j=1}^K \exp \gamma_t S_{j,t}}\right)+\eta||\mathbf{x}_i||_{\mathbf{V}^{-1}_t}
\end{equation}
The gradient estimator $\hat{g}(\beta)$ is obtained by repalcing $\mu_i$ with $\hat{\mu}_{i,t}=\mathbf{x}_i^T\hat{\boldsymbol{\theta}}_t$ where $\hat{\boldsymbol{\theta}}_t=\mathbf{V^{-1}_t}\sum_{s=1}^t \mathbf{x}_sy_s$ is obtained via least-square estimator.

\begin{equation}
    \hat{g}(\beta)=\sum_{t=1}^T \sum_{i=1}^K \hat{\mu}_{i,t} p_{i,t}\left(\gamma_t \phi_{i,t}-\frac{\sum_{j=1}^K \gamma_t \phi_{j,t}\exp \gamma_t S_{j,t}}{\sum_{j=1}^K \exp \gamma_t S_{j,t}}\right)+\eta||\mathbf{x}_i||_{\mathbf{V}^{-1}_t}
\end{equation}

\end{proof}

\section*{Appendix D}
\label{section: appendix_j}
This sextion contains the proof of Theorem~\refC{theorem: lse_soft_regret}.
\begin{proof}
The probability of each arm is defined as 
\begin{equation}
\label{eq: p_i_t_def }
    p_{i,t}=\frac{\exp(\gamma_t S_{i,t})}{\sum_{j=1}^K \exp(\gamma_t S_{j,t})}
\end{equation}
$S_{i,t}$ is defined as
\begin{equation}
\label{eq: s_i_t_def}
    S_{i,t}=\hat{\beta}\phi_{i,t}-\hat{\Delta}_{i,t}=\hat{\beta} (||\mathbf{x}_i||_{\mathbf{V}^{-1}_t}+||\mathbf{x}_{i_*}||_{\mathbf{V}^{-1}_t})-(\hat{\mu}_{i_*,t}-\hat{\mu}_{i,t})
\end{equation}
The cumulative regret to be minimized is defined as 
\begin{equation}
\begin{split}
    R_T=\sum_{t=1}^T \mathbb{E}[r_t]=\sum_{t=1}^T \mu_*-\mathbb{E}[y_t]
    &=\sum_{t=1}^T(\mu_*- \sum_{i=1}^K p_{i,t}\mu_i)\\
    &=\sum_{t=1}^T\sum_{i=1}^K p_{i,t}(\mu_*-\mu_i)=\sum_{t=1}^T\sum_{i=1}^K p_{i,t}\Delta_i
\end{split}
\end{equation}
where we use $\sum_{i=1}^K p_{i,t}=1$.\\
At each time $t$, trm set $\mathcal{A}$ is divided into two subsets $\mathcal{U}_t$ and $\mathcal{L}_t$ with $\mathcal{U}_t\cup \mathcal{L}_t=\mathcal{A}$. Arm $i \in \mathcal{U}_t$ if $S_{i,t}\geq 0$ and arm $i\in \mathcal{L}_t$ if $S_{i,t}<0$. 
\begin{equation}
    \mathbb{E}[r_t]=\sum_{i=1}^K p_{i,t}\Delta_i=\sum_{i\in\mathcal{U}_t}p_{i,t}\Delta_i+\sum_{i \in \mathcal{L}_t}p_{i,t}\Delta_i
\end{equation}
Suppose $\gamma_t$ follows Lemma~\refC{lemma: gamma_lemma}, then $\sum_{i\in \mathcal{L}_t}p_{i,t}< 1-\delta$. Assume $\Delta_i\leq 1,  \forall i\in \mathcal{A}$. Then,
\begin{equation}
    \mathbb{E}[r_t]= \sum_{i\in\mathcal{U}_t}p_{i,t}\Delta_i+\sum_{i \in \mathcal{L}_t}p_{i,t}\leq \sum_{i\in\mathcal{U}_t}p_{i,t}\Delta_i+(1-\delta)
\end{equation}
By setting $\delta \approx 1$, we have $1-\delta \approx 0$. It means arms in $\mathcal{L}_t$ are unlikely to be selected. So, the second term can be dropped. Therefore,
\begin{equation}
    \mathbb{E}[r_t]\leq \sum_{i\in\mathcal{U}_t}p_{i,t}\Delta_i
\end{equation}
Thus, 
\begin{equation}
\begin{split}
\label{eq: r_bound_1}
     \mathbb{E}[r_t]\leq \sum_{i\in\mathcal{U}_t}p_{i,t}\Delta_i
     &=\sum_{i\in \mathcal{U}_t}p_{i,t}(\mu_*-\mu_i)
\end{split}
\end{equation}
Note that at each time $t$, $|\hat{\mu}_{i,t}-\mu_i|\leq \hat{\beta}||\mathbf{x}_i||_{\mathbf{V}^{-1}_t}, \ \forall i \in [K]$. Then 
\begin{equation}
    \mu_*\leq \hat{\mu}_{*,t}+\hat{\beta}||\mathbf{x}_*||_{\mathbf{V}^{-1}_t}
\end{equation}
and 
\begin{equation}
    \mu_i\geq \hat{\mu}_{i,t}-\hat{\beta}||\mathbf{x}_i||_{\mathbf{V}^{-1}_t}
\end{equation}
Thus, 
\begin{equation}
    \mu_*-\mu_i\leq \hat{\beta}(||\mathbf{x}_*||_{\mathbf{V}^{-1}_t}+||\mathbf{x}_i||_{\mathbf{V}^{-1}_t})+(\hat{\mu}_{*,t}-\hat{\mu}_{i,t})
\end{equation}
Note that $\hat{\mu}_{*,t}-\hat{\mu}_{i,t}\leq \hat{\mu}_{i_*,t}-\hat{\mu}_{i,t}$ where $i_*=\arg\max_{j\in [K]}\hat{\mu}_{j,t}-\hat{\mu}_{i,t}$. Therefore, 
\begin{equation}
\begin{split}
     \mu_*-\mu_i
     \leq \hat{\beta}(||\mathbf{x}_*||_{\mathbf{V}^{-1}_t}+||\mathbf{x}_i||_{\mathbf{V}^{-1}_t})+(\hat{\mu}_{i_*,t}-\hat{\mu}_{i,t})
\end{split}
\end{equation}
Since $i \in \mathcal{U}_t$, $S_{i,t}\geq 0$. That is $\hat{\mu}_{i_*,t}-\hat{\mu}_{i,t}\leq \beta(||\mathbf{x}_*||_{\mathbf{V}^{-1}_t}+||\mathbf{x}_i||_{\mathbf{V}^{-1}_t})$. Then, 
\begin{equation}
\begin{split}
    \mu_*-\mu_i
     &\leq \hat{\beta}(||\mathbf{x}_*||_{\mathbf{V}^{-1}_t}+||\mathbf{x}_i||_{\mathbf{V}^{-1}_t})+(\hat{\mu}_{i_*,t}-\hat{\mu}_{i,t})\\
     &\leq 2\hat{\beta} (||\mathbf{x}_*||_{\mathbf{V}^{-1}_t}+||\mathbf{x}_i||_{\mathbf{V}^{-1}_t})
\end{split}
\end{equation}
Define $\psi_t=\max_{i \in [K]}||\mathbf{x}_i||_{\mathbf{V}^{-1}_t}$. We have
\begin{equation}
    \mu_*-\mu_i\leq 4\hat{\beta}\psi_t
\end{equation}
Plugging this into Eq.~\refC{eq: r_bound_1} gives 
\begin{equation}
    \mathbb{E}[r_t]\leq 4\hat{\beta} \sum_{i\in \mathcal{U}_t}p_{i,t}\psi_t
\end{equation}
Since we assume $\gamma_t$ follows Lemma~\refC{lemma: gamma_lemma}, we have $p_{\mathcal{U}_t}=\sum_{i\in \mathcal{U}_t}p_{i,t}=\delta$. Therefore, 
\begin{equation}
    \mathbb{E}[r_t]\leq 4\hat{\beta} \sum_{i\in \mathcal{U}_t}p_{i,t}\psi_t=4\hat{\beta}\phi_t\sum_{i\in \mathcal{U}_t}p_{i,t}=4\hat{\beta}\psi_t p_{\mathcal{U}_t}\leq 4\hat{\beta}\delta \psi_t
\end{equation}
Thus, the cumulative regret
\begin{equation}
\label{eq: R_1}
\begin{split}
    R_T=\sum_{t=1}^T \mathbb{E}[r_t]\leq \sqrt{T\sum_{t=1}^T \mathbb{E}[r_t]^2}\leq 4\hat{\beta} \delta \sqrt{T\sum_{t=1}^T\psi_t^2}
\end{split}
\end{equation}
From Lemma~\refC{lemma: sum_x} (stated below), we have
\begin{equation}
    \sum_{t=1}^T\psi_t^2\leq 2d\log(\alpha+\frac{T}{d})
\end{equation}
Plugging in Eq.~\refC{eq: R_1}, 
\begin{equation}
    R_T\leq 4\hat{\beta} \delta \sqrt{2Td\log(\alpha+\frac{T}{d})}=\tilde{\mathcal{O}}(\hat{\beta}\sqrt{Td\log(1+\frac{T}{d})})
\end{equation}
where $\delta$ is the probability parameter chosen by user.

\begin{lemma}
\label{lemma: sum_x}
(Lemma 11 in \Ccite{abbasi2011improved})
\begin{equation}
    \sum_{t=1}^{T} ||\mathbf{x}||^2_{\mathbf{V}^{-1}_t}\leq \log det(\mathbf{V}_t)\leq 2d\log(\alpha+\frac{T}{d})
\end{equation}
\end{lemma}

\end{proof}

\newpage
\section*{Appendix E}
This section contains the pseudo code of \algo{SoftUCB}, \algo{SoftUCB offline} and \algo{SoftUCB online}.

\begin{algorithm}[H]
\SetKwInOut{Input}{Input}\SetKwInOut{Output}{Output}
\Input{$\beta$, $\mathcal{A}$, $K$, $T$, $\alpha$. }
\BlankLine
\textbf{Initialization~~~:} $\mathbf{V}_0=\alpha \mathbf{I}\in \mathbb{R}^{d\times d}$, $\mathbf{b}_0=\mathbf{0}\in \mathbb{R}^d$, $\hat{\boldsymbol{\theta}}_0=\mathbf{0}\in \mathbb{R}^d$,  $\gamma_0=0$.\\
\For{$t \in [1, T]$}{
\begin{enumerate}
    \item Find $S_{i,t}, \forall i\in \mathcal{A}$ via Eq.~\refC{eq: s_i} with $\beta$.
    \item Find $\boldsymbol{\pi}_t$ via Eq.~\refC{eq: p_i} with $\gamma_{t-1}$.
    \item Select arm $i_t \in \mathcal{A}$ randomly following $\boldsymbol{\pi}_t$ and receive payoff $y_t$.
    \item Update $\mathbf{V}_t\gets \mathbf{V}_{t}+\mathbf{x}_t\mathbf{x}_t^T$, $\mathbf{b}_t\gets \mathbf{b}_{t-1}+\mathbf{x}_ty_t$ and $\hat{\boldsymbol{\theta}}_t=\mathbf{V}_{t}^{-1}\mathbf{b}_{t}$.
    \item Update $\gamma_t$ via Eq.~\refC{eq: emp_gamma} .
\end{enumerate}
}
\caption{\algo{SoftUCB}}
\label{algorithm: exp_ucb}
\end{algorithm}

\begin{algorithm}[H]
\SetKwInOut{Input}{Input}\SetKwInOut{Output}{Output}
\Input{$\mathcal{A}$, $K$, $T$,  $\lambda$, $\eta$}
\BlankLine
\textbf{Initialization~~~:} $\beta_0=0$, $\hat{\beta}=0$.\\
\For{$n \in [1, N]$}{
\begin{enumerate}
\item Run \algo{SoftUCB} on $\mathcal{A}$ rounds with $\beta=\beta_{n-1}$.
\item Update $\beta_{n}\gets \beta_{n-1}+\lambda\hat{g}(\beta)$ via Eq.~\refC{eq: emp_beta_g}
\end{enumerate}
}
\Output{$\hat{\beta} \gets \beta_N$}
\BlankLine
Run \algo{SoftUCB} on $\mathcal{A}$ with $\beta=\hat{\beta}$.
\caption{\algo{SoftUCB offline}}
\label{algorithm: exp_ucb_offline}
\end{algorithm}

\begin{algorithm}[H]
\SetKwInOut{Input}{Input}\SetKwInOut{Output}{Output}
\Input{$\mathcal{A}$, $K$, $T$, $\alpha$, $\lambda$, $\eta$}
\BlankLine
\textbf{Initialization~~~:} $\beta_0=0$, $\mathbf{V}_0=\alpha I\in \mathbb{R}^{d\times d}$, $\mathbf{b}_0=\mathbf{0}\in \mathbb{R}^d$, $\hat{\boldsymbol{\theta}}_0=\mathbf{0}\in \mathbb{R}^d$,  $\gamma_0=0$.\\
\For{$t \in [1, T]$}{
\begin{enumerate}
    \item Select arm $i_t\in [K]$ randomly following $\boldsymbol{\pi}_t$ and receive payoff $y_t$.
    \item Update $\mathbf{V}_t\gets \mathbf{V}_{t}+\mathbf{x}_t\mathbf{x}_{t}^T$, $\mathbf{b}_t\gets \mathbf{b}_{t-1}+\mathbf{x}_{t}\mathbf{y}_t$ and $\hat{\boldsymbol{\theta}}_t=\mathbf{V}_{t}^{-1}\mathbf{b}_{t}$.
    \item Update $\beta_t\gets \beta_{t-1}+\lambda \hat{g}_t(\beta)$ via Eq.~\refC{eq: emp_beta_g_online}.

\end{enumerate}
}
\caption{\algo{SoftUCB online}}
\label{algorithm: exp_ucb_online}
\end{algorithm}

\newpage
\section*{Appendix F}
This section contains the learning curves of \algo{SoftUCB offline}.

\begin{figure}[H]
    \centering
    \begin{subfigure}{0.24\textwidth}
        \includegraphics[width=\textwidth, height=4cm]{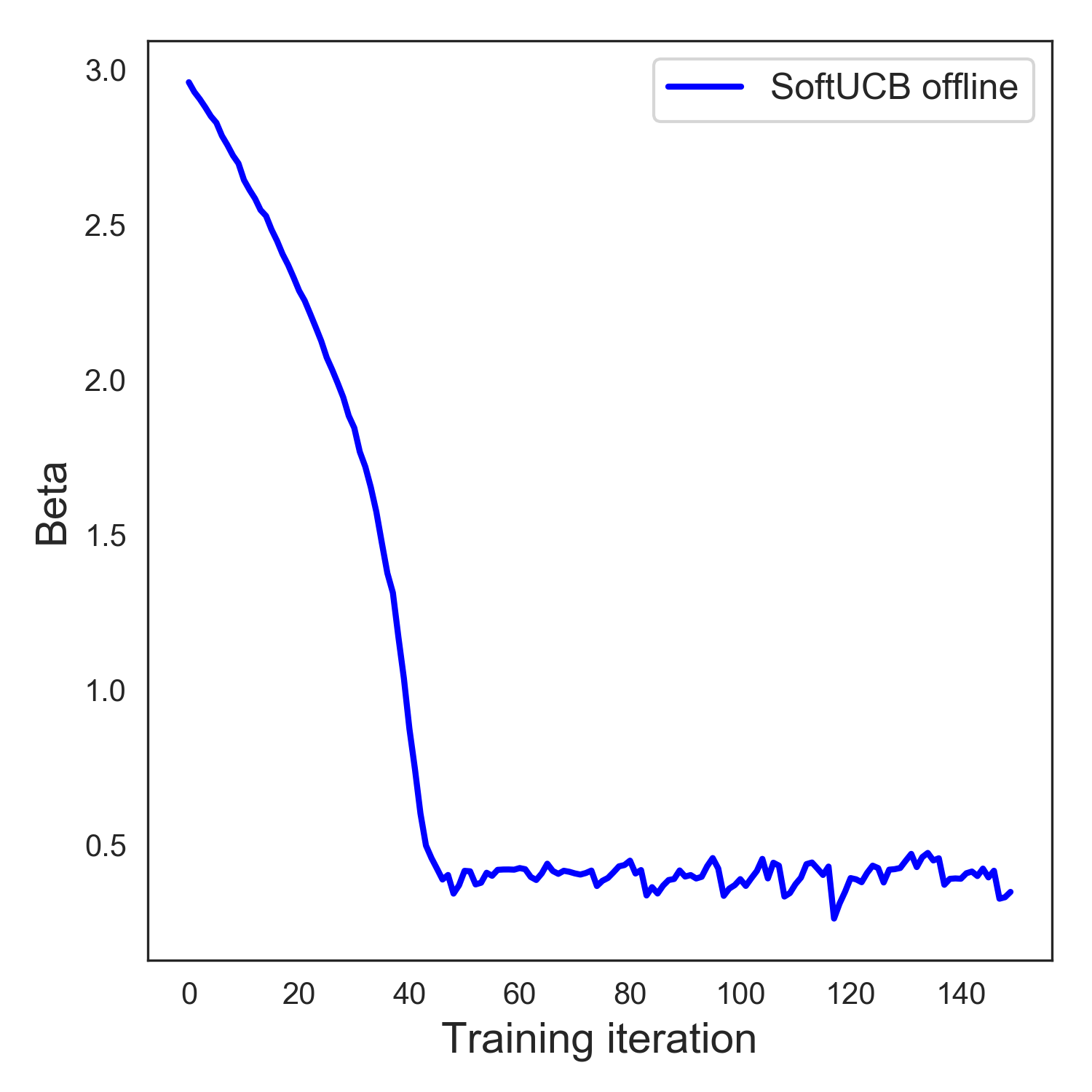}
        \subcaption{$d=5, T=2^8$}
    \end{subfigure}
    \begin{subfigure}{0.24\textwidth}
        \includegraphics[width=\textwidth, height=4cm]{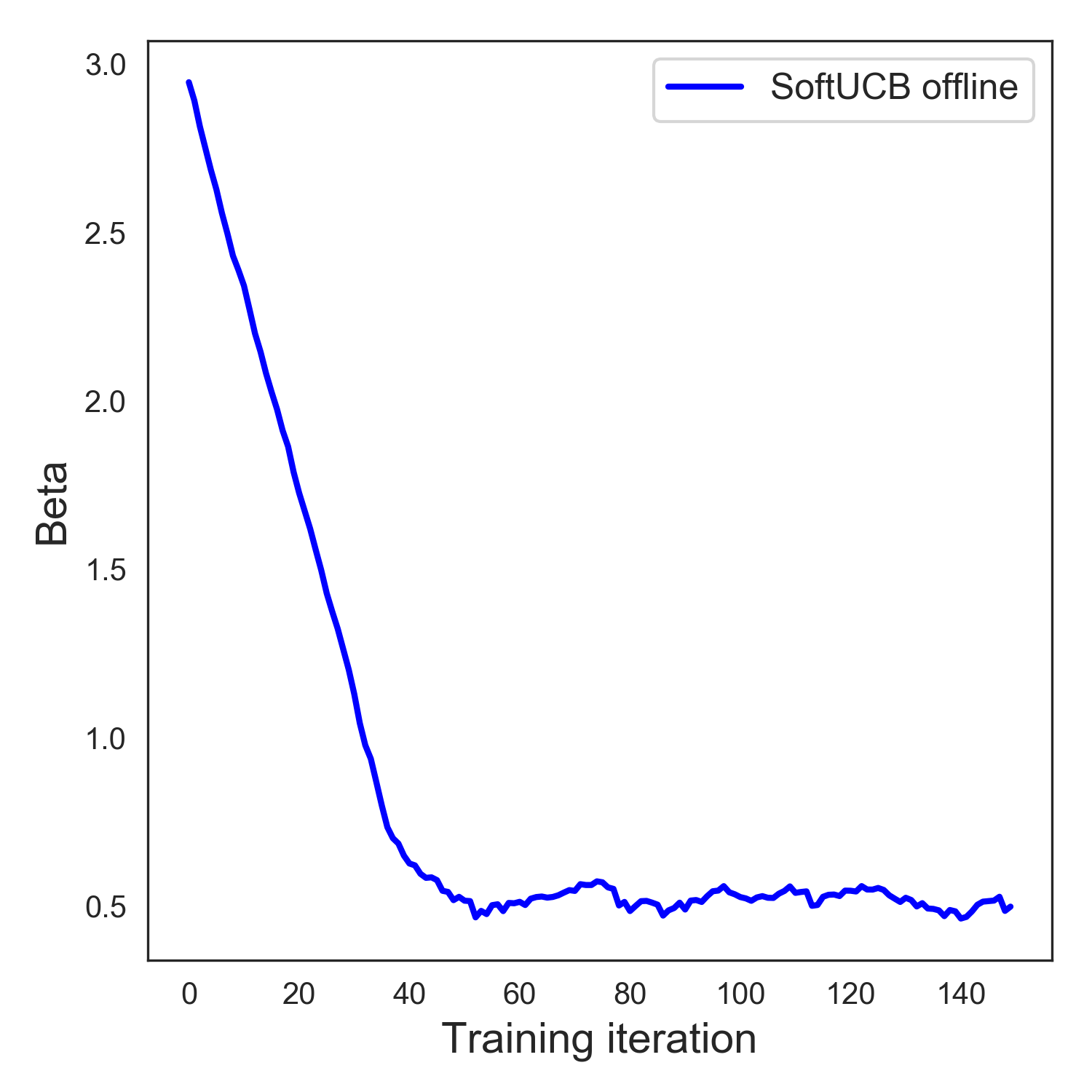}
        \subcaption{$d=5, T=2^9$}
    \end{subfigure}
    \begin{subfigure}{0.24\textwidth}
        \includegraphics[width=\textwidth, height=4cm]{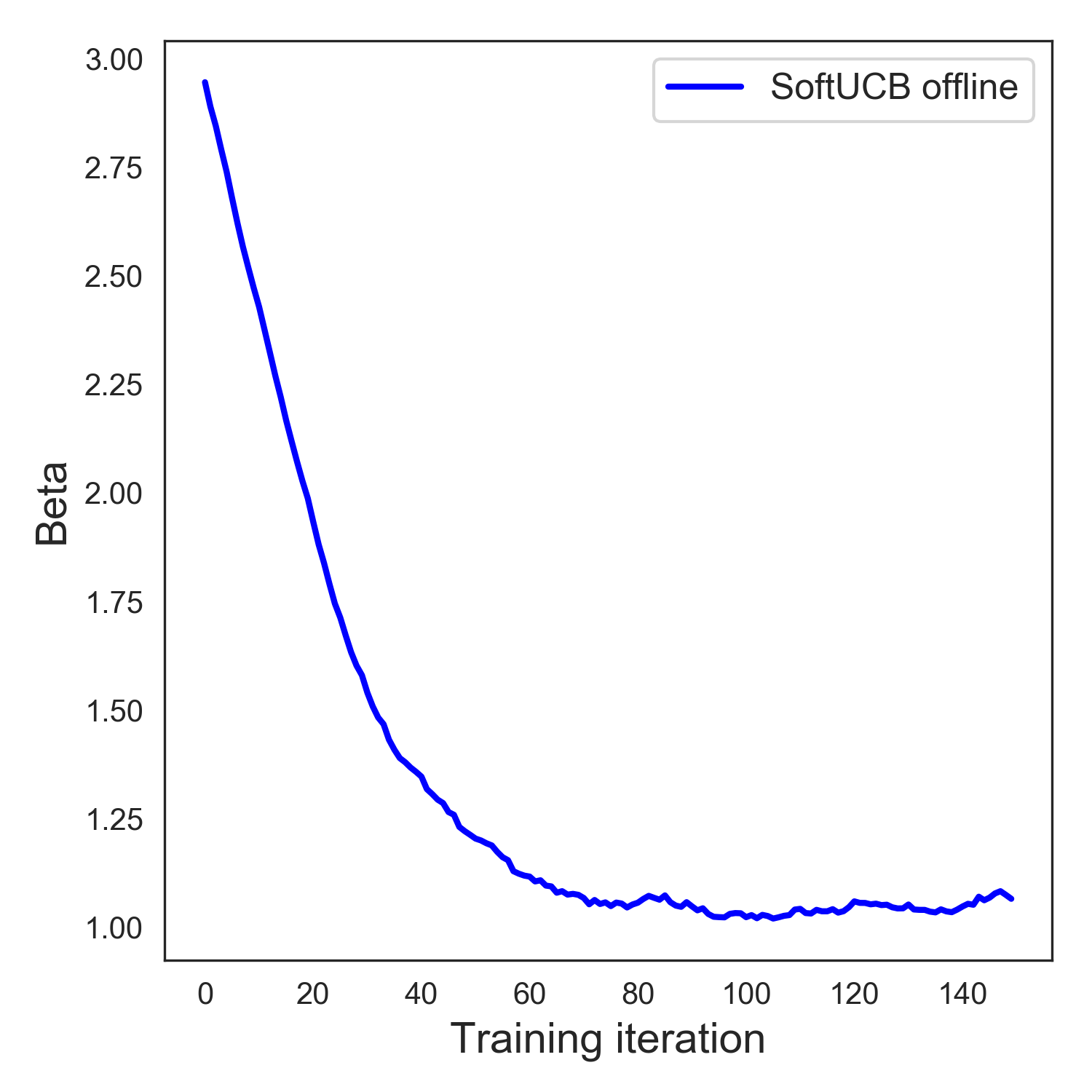}
        \subcaption{$d=5, T=2^{10}$}
    \end{subfigure}
\caption{Learning curves of \algo{SoftUCB offline}}
\label{fig: learn_curves_offline_first_part}
\end{figure}

\begin{figure}[H]
    \centering
    \begin{subfigure}{0.24\textwidth}
        \includegraphics[width=\textwidth, height=4cm]{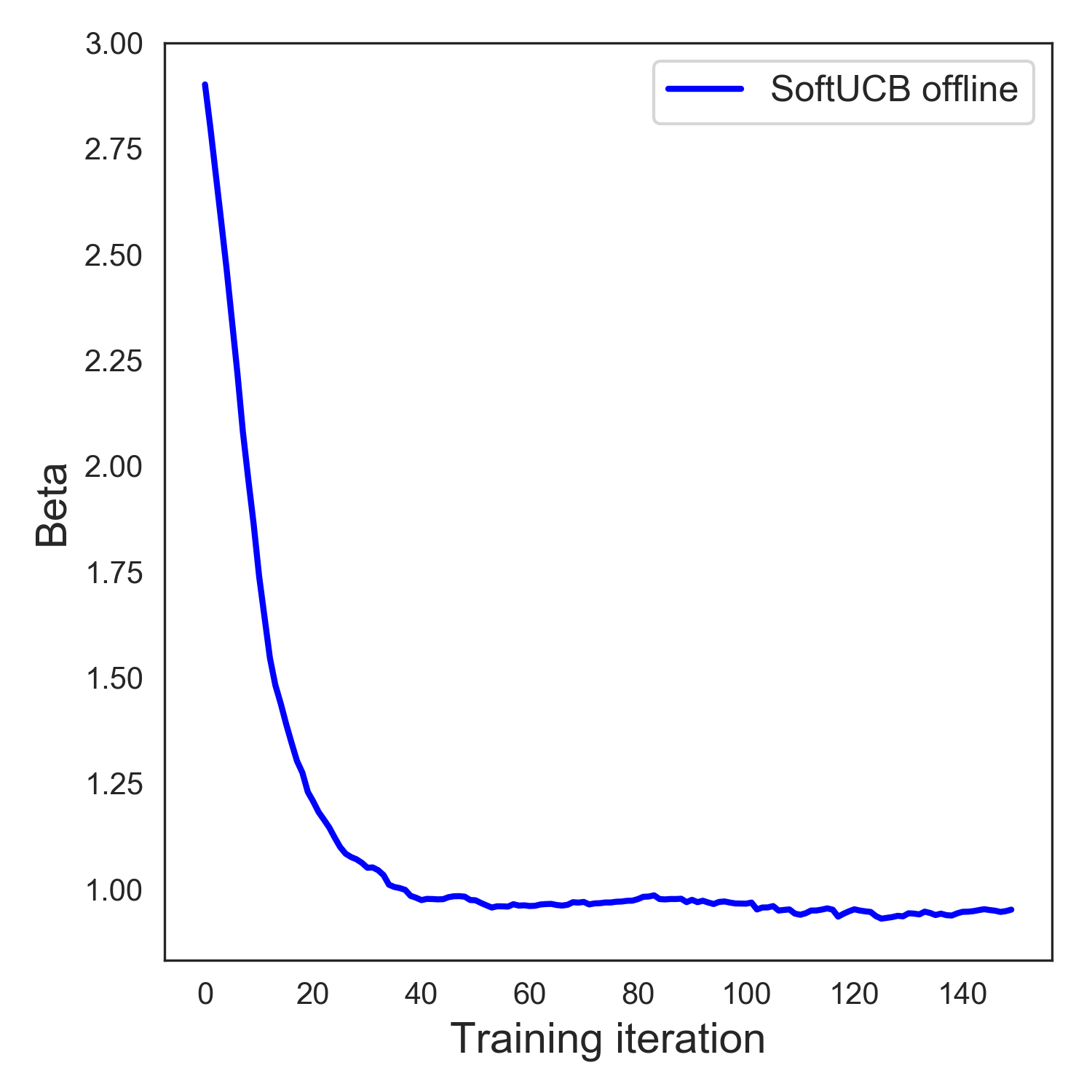}
        \subcaption{$d=10, T=2^{10}$}
    \end{subfigure}
    \begin{subfigure}{0.24\textwidth}
        \includegraphics[width=\textwidth, height=4cm]{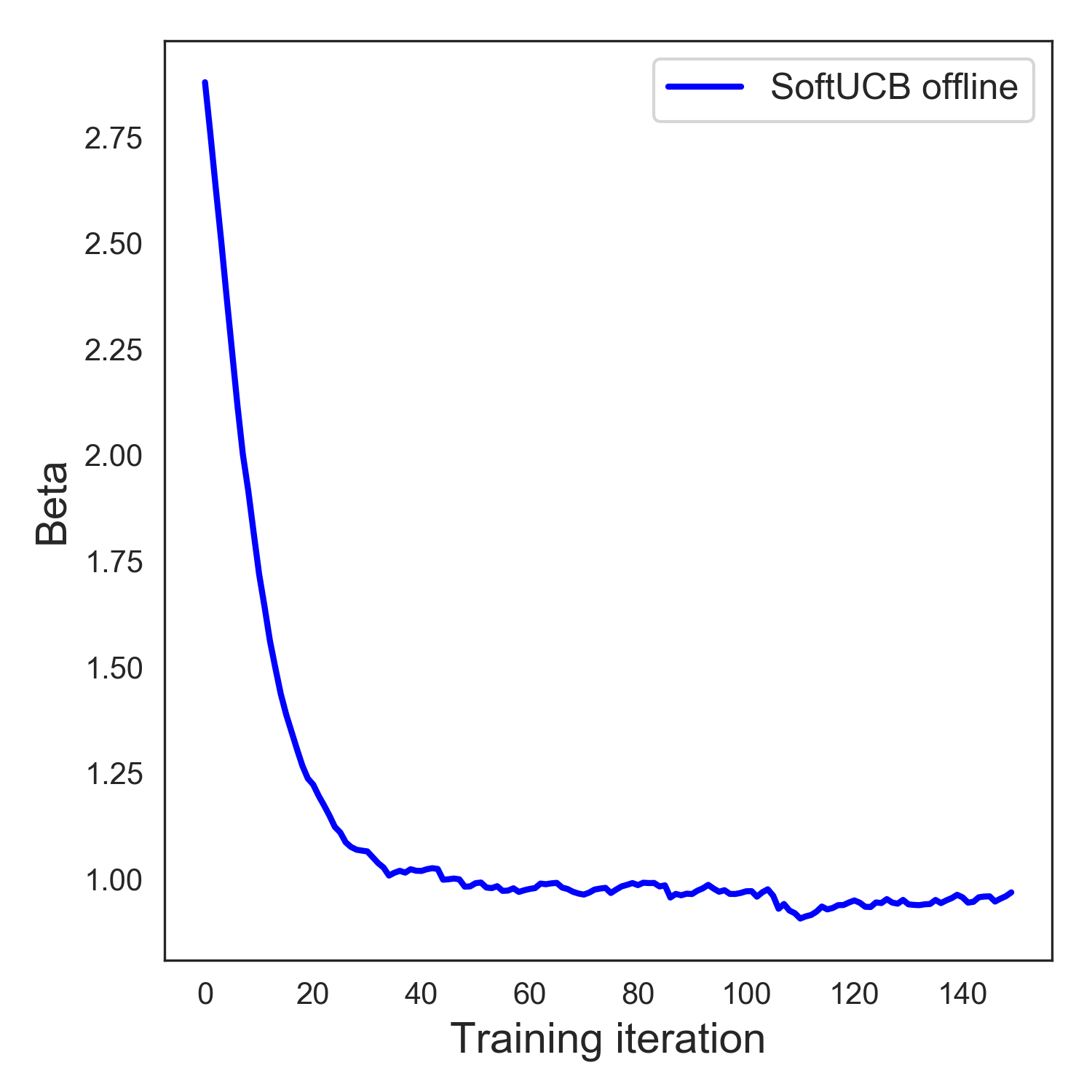}
        \subcaption{$d=15, T=2^{10}$}
    \end{subfigure}
\caption{Learning curves of \algo{SoftUCB offline}}
\label{fig: learn_curves_offline_second_part}
\end{figure}

\section*{Appendix G}
The dataset \textbf{Jester} contains ratings of 40 jokes from 19891 users. We sample $K=50$ users randomly as arms. Their rating to top 39 jokes are used as feature vector. Then, to reduce the sparsity, we apply principle component analysis algorithm to reduce the dimension $d=10$. Their rating on the 40th jokes are used as rewards.  At each round, the algorithm selects on user to recommend the joke and the reward is the rating given by the user. \textbf{MovieLens} contains 6k users and their ratings on 40k movies. Since not every user gives ratings on all movies, there are a large mount of missing ratings. We factorize the rating matrix to fill the missing values. The rest works the same as in \textbf{Jester}.

\end{document}